%% file: norddrg-benchmark-arxiv.tex
\documentclass{article}

\usepackage{arxiv}

\usepackage[utf8]{inputenc} 
\usepackage[T1]{fontenc}    
\usepackage{hyperref}       
\usepackage{url}            
\usepackage{booktabs}       
\usepackage{amsfonts}       
\usepackage{nicefrac}       
\usepackage{microtype}      
\usepackage{lipsum}
\usepackage{graphicx}

\usepackage{algorithm2e}

\usepackage{booktabs}   
\usepackage{makecell}   

\usepackage{tabularx}  
\usepackage{booktabs}  
\usepackage{array}     

\usepackage{amsthm}

\theoremstyle{definition}

\usepackage{epsfig}
\usepackage{subcaption}
\usepackage{calc}
\usepackage{amssymb}
\usepackage{amstext}
\usepackage{amsmath}
\usepackage{amsthm}
\usepackage{multicol}
\usepackage{pslatex}
\usepackage{apalike}
\usepackage{algorithm2e}
\usepackage[bottom]{footmisc}

\usepackage{enumitem}   
\usepackage{hyperref}   

\usepackage{booktabs}   
\usepackage{tabularx}   
\usepackage{array}     
\usepackage{booktabs}  

\usepackage{multirow} 

\usepackage{xcolor}      
\usepackage{pifont}      
\newcommand{\cmark}{{\color{green!60!black}\ding{51}}} 
\newcommand{\xmark}{{\color{red!70!black}\ding{55}}}   

\graphicspath{ {./images/} }

\title{NordDRG AI Benchmark for Large Language Models}

\author{
  Tapio Pitkäranta \\
  Department of Computer Science and Engineering\\
  Aalto University, 
  Finland \\
  \texttt{tapio.pitkaranta@iki.fi} \\
}

\begin{document}
\maketitle

\input{norddrg-benchmark-abstract.tex}

\keywords{Large Language Models (LLMs), LLM Software Agents, Diagnosis-Related Groups (DRG), NordDRG, Clinical coding, Healthcare reimbursement, CaseMix systems, Benchmarking and reproducibility, Grouper emulation}

\input{norddrg-benchmark-sections.tex}

\bibliographystyle{apalike}
{\small
\bibliography{../../references,references_curated_short,../../references_hada_short}}


\end{document}

%% file: norddrg-benchmark-abstract.tex
\begin{abstract}
  \textbf{Problem.} Large language models (LLMs) are being piloted for clinical coding and decision support, yet no open benchmark targets the \emph{hospital-funding} layer where Diagnosis-Related Groups (DRGs) determine reimbursement. In most OECD systems, DRGs route a substantial share of multi-trillion-dollar health spending through governed grouper software, making transparency and auditability first-order concerns rather than mere implementation details.

  \textbf{Contribution.} We release \emph{NordDRG-AI-Benchmark}, the first public, \emph{rule-complete} test-bed for DRG reasoning. It bundles \textbf{(i)} machine-readable \(\sim\!20\)-sheet NordDRG definition tables and \textbf{(ii)} expert manuals and change-log templates that capture governance workflows, and exposes two suites: a 13-task \emph{Logic} benchmark (code lookup, cross-table inference, grouping features, multilingual terminology, and CC/MCC validity checks) and a 13-task \emph{Grouper} benchmark that requires \emph{full DRG-grouper emulation} with strict exact-match scoring on \emph{both} the DRG and the triggering \texttt{drg\_logic.id}. Lightweight reference agents (LogicAgent, GrouperAgent) enable artefact-only evaluation.

  \textbf{Results.} Under an artefact-only (no-web) setting, on the 13 Logic tasks \emph{GPT-5 Thinking} and \emph{Opus 4.1} score \textbf{13/13}, \emph{o3} \textbf{12/13}; mid-tier models (\emph{GPT-5 Thinking Mini}, \emph{o4-mini}, \emph{GPT-5 Fast}) achieve \textbf{6–8/13}, and the remaining models score \textbf{5/13 or below}. On full grouper emulation across 13 tasks (exact match on \emph{both} DRG and \texttt{drg\_logic.id}), \emph{GPT-5 Thinking} solves \textbf{7/13}, \emph{o3} \textbf{6/13}, and \emph{o4-mini} \textbf{3/13}; \emph{GPT-5 Thinking Mini} solves \textbf{1/13}, and all other tested endpoints score \textbf{0/13}. To our knowledge, this is the first public report of an LLM \emph{partially emulating} the complete NordDRG grouper logic with governance-grade traceability.

  \textbf{Significance.} Coupling a rule-complete release with exact-match tasks and open scoring reveals domain-specific strengths and weaknesses that generic leaderboards miss and provides a reproducible yardstick for head-to-head and longitudinal evaluation in hospital funding.

  \textbf{Availability.} All artefacts and scripts are available at \url{https://github.com/longshoreforrest/norddrg-ai-benchmark}.
\end{abstract}

%% file: norddrg-benchmark-sections.tex

\section{Introduction}

\subsection{Problem Identification} \label{sec:problem-identification}
Large-scale generative models have progressed from research prototypes to production-grade services in under five years, bringing natural-language interfaces to domains once considered too specialised for automated reasoning. Healthcare finance is a notable example: hospital payments in many health systems are determined by \emph{case-based} schemes built on Diagnosis-Related Groups (DRGs). In Europe, DRG systems have become the de facto basis for paying hospitals across most industrialised countries; in the United States, Medicare’s Inpatient Prospective Payment System (IPPS) pays acute inpatient stays by MS-DRG; and across OECD members, DRG-like payments are among the most common approaches to reimbursing acute care. These arrangements channel a substantial share of public hospital expenditure through DRG groupers, making their transparency and auditability a matter of fiscal governance as well as clinical accountability \cite{Busse2011,WHO2020DRG,OECD2023PaymentModels,CMSIPPS}.

\paragraph{Scale and stakes.}
Global current health expenditure reached about US\$9.8~trillion in 2021 (10.3\% of world GDP) \cite{WHO2024GlobalSpending}. Hospitals absorb the largest share of this spending—around one third worldwide (35.4\% in 2017) \cite{Schneider2021Providers} and roughly 39\% across OECD countries \cite{OECD2023HAG}. In systems that use DRGs, a sizeable portion of these flows is mediated by grouper software, underscoring the need for auditable, rule-faithful logic.

\paragraph{Decision-model framing.}
A DRG grouper is, in effect, a \emph{complex automated decision-making model} that maps a patient episode’s \emph{minimum dataset}—principal diagnosis, secondary diagnoses (with CC/MCC implications), procedure/intervention codes, age and sex, and context such as MDC entry and care setting—onto a single payment group via an ordered set of rules with explicit inclusions, exclusions, and national activation flags. Fiscal stakes arise because this deterministic mapping selects the payment from structured inputs; governance stakes arise because each decision should be traceable to specific rule rows and their change history.

\paragraph{Measurement gap.}
Yet the rule sets that govern DRG allocations are intricate, multi-table specifications revised annually and interpreted through expert manuals and change-control workflows. For non-experts, these systems remain opaque; for experts, verifying end-to-end logic and its governance trace is laborious. Despite a surge in medical-LLM research, there has been \emph{no open, rule-complete benchmark} to assess whether models can read the DRG rule graph and reproduce the grouper’s governed behaviour. Prior evaluations either flatten the task into single-label classification on proprietary data or probe isolated rules, leaving no operational yardstick for measuring progress on hospital-funding logic. This gap motivates the benchmark introduced in this article.

\subsection{Contributions}
We release \emph{NordDRG-AI-Benchmark}, a rule-complete, publicly available testbed that unites clinical coding, multilingual reasoning, and healthcare finance. The contribution has five parts:

\begin{enumerate}[nosep,leftmargin=1.6em]
  \item \textbf{Definition tables and expert manuals (A1–A2).}
        (i) \emph{definition tables} \textit{(Section~\ref{norddrg-tables})} comprising $\sim\!20$ inter-linked sheets that encode the full NordDRG grouper logic (diagnoses/procedures, properties, country activation), and (ii) \emph{expert manuals} \textit{(Section~\ref{norddrg-docs})} and change-log templates that document the real governance workflow. Lightweight FI and FI–EN excerpts and ready-made dataset configurations are provided for rapid prototyping (cf.\ Section~\ref{norddrg-fi}--\ref{norddrg-dataset-configurations}).

  \item \textbf{Two benchmark suites with exact-match scoring.}
        \emph{NordDRG Logic Benchmark} (13 CaseMix tasks, Section~\ref{norddrg-questions}) probes rule-level skills—code lookup, cross-table joins, grouping features, multilingual terminology, and QA audits. 
        \emph{NordDRG Grouper Benchmark} (13 case-level grouping tasks, Section~\ref{norddrg-benchmark-grouper}) requires full emulation of the specification’s control flow over \texttt{drg\_logic} (execution order \texttt{ORD}, age/sex bounds, MDC entry, OR/PROCPRO evidence, CC/MCC with exclusions, national activation) and returns both \texttt{drg\_nat} and the triggering \texttt{drg\_logic.id}. 
        Gold answer keys enable strict, governance-grade evaluation.

  \item \textbf{Empirical baselines.}
        Under a no-web, artefact-constrained setting, top models solve all 13 \emph{Logic} tasks (GPT-5 Thinking, Opus~4.1: 13/13; o3: 12/13), whereas full grouper emulation remains challenging (best to date: 7/13 exact matches on \emph{Grouper-1--13}); see Section~\ref{sec:benchmark-demo}.

  \item \textbf{Lightweight reference agents.} Provider-agnostic \emph{LogicAgent} (Combined tables) and \emph{GrouperAgent} (FI tables) for artefact-only evaluation; details in Section~\ref{subsec:lightweight-agents}.

  \item \textbf{Availability and reproducibility.}
        All artefacts, prompts, and gold answers are released in a versioned public repository with schema-stable identifiers and drop-in annual updates (Sections~\ref{sec:design-development} and \ref{sec:conclusions}).
\end{enumerate}

\subsection{Research Questions}\label{sec:research-questions}
We instantiate the problem statement into three research questions that map directly to our design–science objectives:

\begin{enumerate}[nosep,leftmargin=1.4em]
  \item \textbf{RQ1 — Artefact design (O1).} What \emph{minimal, versioned} artefact bundle and schema are necessary and sufficient to encode the complete NordDRG rule graph and its governance semantics in a public, machine-readable form?

  \item \textbf{RQ2 — Benchmark operationalisation (O2).} How should tasks, inputs, and exact-match scoring be structured so that (i) the \emph{Logic} suite measures rule-level reasoning over the definition tables and manuals, and (ii) the \emph{Grouper} suite enforces \emph{full emulation} of the official control flow—returning both \texttt{drg\_nat} and the triggering \texttt{drg\_logic.id}?

  \item \textbf{RQ3 — Baseline capability and failure modes (O3).} When constrained to the released artefacts (no external web access), how do representative LLM endpoints perform on the \emph{Logic} and \emph{Grouper} suites, and which systematic errors account for misses (e.g., cross-table joins, \texttt{ORD} priority, CC/MCC exclusions, national activation)?
\end{enumerate}

\noindent These questions align one-to-one with the objectives in Section~\ref{sec:research-objectives} and are answered empirically in Section~\ref{sec:benchmark-demo}.

\subsection{Structure of the paper}\label{sec:structure-of-the-paper}
The remainder of the paper is organised as follows.
Section~\ref{sec:background-related} positions our contribution within the broader literature.
Section~\ref{sec:benchmark-research-methodology} sets out the research methodology, gap, and objectives.
Section~\ref{sec:design-development} details the benchmark's design and development.
Section~\ref{sec:benchmark-demo} reports baseline LLM performance and assesses coverage, extensibility, and differentiation.
Section~\ref{sec:conclusions} concludes with availability, repository access, and contribution guidance.

\section{Background and Related Work} \label{sec:background-related}

\noindent To orient the reader:
Section~\ref{sec:norddrg-casemix} introduces the NordDRG CaseMix system—the rule base our benchmark operationalises—and establishes domain context.
Section~\ref{sec:llms} reviews core advances in LLMs relevant to our tasks.
Section~\ref{subsec:benchmark-background-agents} outlines lightweight LLM software agents as a framing for the benchmark.
Section~\ref{subsec:benchmark-alignment} motivates governance-grade evaluation through alignment and safety considerations.
Section~\ref{sec:benchmark-llm-leaderboards} surveys general LLM benchmarks and leaderboards, clarifying what they measure—and what they miss for CaseMix reasoning.
Section~\ref{sec:related-drg-llm} synthesises prior DRG-LLM studies to surface the remaining gaps.
Section~\ref{subsec:drg-grouper-software} describes production DRG grouper software and situates our benchmark within that ecosystem.
Section~\ref{sec:our-previous-work} differentiates this article from our earlier publications by specifying reused versus novel contributions.

\subsection{NordDRG CaseMix System} \label{sec:norddrg-casemix}

The Nordic Diagnosis-Related Groups (NordDRG) classification is the shared  
CaseMix framework used throughout the Nordic countries to cluster inpatient  
episodes with comparable resource consumption \cite{nordic_casemix_centre}.  
By assigning each hospital stay to a single DRG, the system underpins  
transparent benchmarking of costs, outcomes, and efficiency across  
institutions, regions, and years.  

NordDRG is distributed as a set of \emph{definition tables}—about twenty  
interlinked sheets released annually in Excel, CSV, and JSON formats—that  
encode the logic connecting diagnosis and procedure codes, age/sex splits, 
and national activation flags to individual DRG codes.  
These tables constitute the authoritative rule base for coding software,  
technical-change management, and policy analysis.  

Because each rule specifies explicit inclusion criteria, identical cases are  
grouped consistently regardless of hospital or country.  At the same time,  
each nation can activate, deactivate, or re-weight DRGs locally, allowing the  
common core to accommodate diverse clinical practices and tariff structures.  
This blend of rule-level precision and country-level flexibility makes NordDRG  
a durable instrument for hospital finance, clinical research, and health-system  
governance in the Nordic region.

\paragraph{Bridge to our contribution.}
The operational details described here are the substrate our benchmark formalises: we package the rule base and governance materials into machine-readable artefacts and task prompts. This lets models be tested on the same criteria human coders and policy analysts use, enabling exact, reproducible checks against the NordDRG logic.

\subsection{Large Language Models (LLMs)} \label{sec:llms}

The breakthrough Transformer architecture introduced by
Vaswani~\textit{et al.}~\cite{vaswani2017attention} ignited today’s
natural–language processing renaissance.  Initial encoder models such
as BERT~\cite{devlin2018bert} were quickly followed by vast
autoregressive generators—GPT-3~\cite{brown2020language}, GPT-4, and
others—whose parameter counts now reach into the hundreds of billions
and display emergent strengths in generation, retrieval, and
reasoning.  Concurrent strands of work explored alternative training
frameworks (T5’s text-to-text paradigm~\cite{raffel2020t5}),
compute–data scaling laws (Chinchilla~\cite{hoffmann2022chinchilla}),
and extreme model sizes, exemplified by PaLM’s
540-billion-parameter Pathways network~\cite{chowdhery2022palm}.
Meanwhile, open-weight initiatives such as OPT~\cite{zhang2022opt} and
Llama-2~\cite{touvron2023llama2} have broadened community access to
state-of-the-art capabilities.  Notably, recent evaluations claim that
some LLMs now satisfy modified Turing-test criteria
\cite{jones2025largelanguagemodelspass}, underscoring their increasing
facility for human-like dialogue.  Comprehensive surveys map this
rapid trajectory from early generative models to ChatGPT, situating LLMs within the wider
generative-AI landscape and cataloguing key architectural and
application trends
\cite{naveed2023comprehensive,cao2023comprehensive}.

Since late 2022, chat-style interfaces—epitomised by ChatGPT—have
moved LLMs beyond research into mainstream workflows.  
By masking low-level prompts behind natural conversation,
these interfaces transform LLMs into a versatile cognitive layer that
clinicians, lawyers, educators, and business users can harness with
minimal technical overhead.

\paragraph{Bridge to our contribution.}
Given LLMs’ strengths and limits in reasoning, retrieval, and multilingual handling, our benchmark deliberately mixes cross-table joins, minority-language descriptors, and governance reasoning tasks. This design probes capabilities that generic NLP suites seldom touch, producing signals directly relevant to CaseMix workflows.

\subsection{Emerging LLM Software Agents}\label{subsec:benchmark-background-agents}
\emph{LLM software agents}—systems that pair a large language model with an explicit control loop—expose three core elements: (i) a prompt or system specification encoding the initial policy, (ii) a mutable interaction trace functioning as state, and (iii) the ability to \texttt{act} via tools or APIs \cite{cemri2025multiagentllmsystemsfail,naveed2024comprehensiveoverviewlargelanguage}. Built on the LLM’s perception and inference abilities, such agents support natural-language understanding and generation, task decomposition and planning, decision making, and tool use \cite{xi2023rise,wang2023survey,yang2023foundation}. In practice, a user request can be broken into sub-tasks, relevant information fetched or computed through tools, and the outcome aggregated into a final response \cite{yao2022react,shinn2023reflexion}.

Despite rapid progress, challenges remain in high-stakes domains: hallucinations and brittle reasoning motivate grounding and retrieval-augmented generation (RAG) \cite{ji2023survey,lewis2020retrieval}; safety, ethical risks, and inherited biases require guardrails \cite{bender2021dangers,weidinger2021ethical}; and opaque internal reasoning complicates debugging and audit \cite{doshi2017towards}. Current work targets higher factuality, better explainability, and more reliable coordination (e.g., web-grounded training \cite{nakano2021webgpt}, dynamic context selection \cite{ram2023context}) alongside richer evaluation settings.

\paragraph{Bridge to our work.}
In this article, the agentic perspective is used pragmatically: lightweight, \emph{single-role} agents can wrap deterministic table operations over the released artefacts. A \emph{LogicAgent} can execute rule-level lookups and joins for the Logic tasks, emitting order-invariant, de-duplicated code sets; a \emph{GrouperAgent} can apply the official \texttt{drg\_logic} control flow (including \texttt{ORD}, age/sex bounds, OR/PROCPRO, CC/MCC with exclusions, and national activation) and return both \texttt{drg\_nat} and the triggering \texttt{drg\_logic.id}. This lightweight framing aligns with CaseMix governance by constraining tools to the released tables/PDFs and producing auditable traces, without requiring a full multi-agent pipeline (cf.\ Section~\ref{subsec:lightweight-agents}).

\subsection{AI alignment and safety} \label{subsec:benchmark-alignment}

As technical frontiers expand, the focus of responsible AI practice is shifting from \emph{capability} to \emph{governance}. Ensuring that AI model behaviour aligns with human values and organisational objectives is a critical—and multidimensional—challenge \cite{christian2020alignment}, with well-documented risks of opacity, disparate impact, and governance failures in high-stakes domains \cite{o2017weapons}. The alignment problem is often decomposed into \emph{outer alignment} (does the specified objective capture what we actually want?) and \emph{inner alignment} (does the learned optimiser in fact pursue that objective, especially on inputs outside the training distribution?) \cite{hubinger2019risks,shah2022goal}. Practical failures arise from reward misspecification, specification gaming, and goal misgeneralisation, particularly under non-stationarity \cite{amodei2016concrete,shah2022goal}.

Recent advances combine data-driven optimisation with human oversight to better shape training objectives. Preference learning and reward modelling translate human (or AI-assisted) judgments into training signals \cite{christiano2017deep,leike2018scalable}; instruction tuning with human or AI feedback (e.g., RLHF and constitutional prompting) further reduces harmful behaviours and improves adherence to stated guidelines \cite{ouyang2022training,bai2022constitutional}. However, these techniques alone do not satisfy governance-grade requirements in institutional settings, where decision logic audit trail, deliberation, and auditability for consequential changes are indispensable.

\paragraph{Bridge to our contribution.}
Auditability and governance needs motivate our emphasis on determinism, traceability, and documentation: tasks come with exact-match answer keys and all artefacts are versioned. This structure supports accountable evaluation and makes failures inspectable by domain experts.

\subsection{LLM Benchmarks and Leaderboards} \label{sec:benchmark-llm-leaderboards}

Benchmark suites and public leaderboards are central to tracking progress in large language models. Early collections such as GLUE\,\cite{wang2018glue} and SuperGLUE\,\cite{wang2019superglue} gave way to broader, multi-skill batteries—e.g., BIG-bench\,\cite{bigbench2022}, MMLU\,\cite{hendrycks2021mmlu}, and BabyLM\,\cite{warstadt2023babylm}—probing factual knowledge, reasoning, multilinguality, and instruction following. Meta-evaluations like HELM\,\cite{liang2022helm} further standardise conditions and report robustness, bias, and efficiency alongside accuracy.

\begin{figure}[!h]
  \centering
  \includegraphics[width=0.7\textwidth]{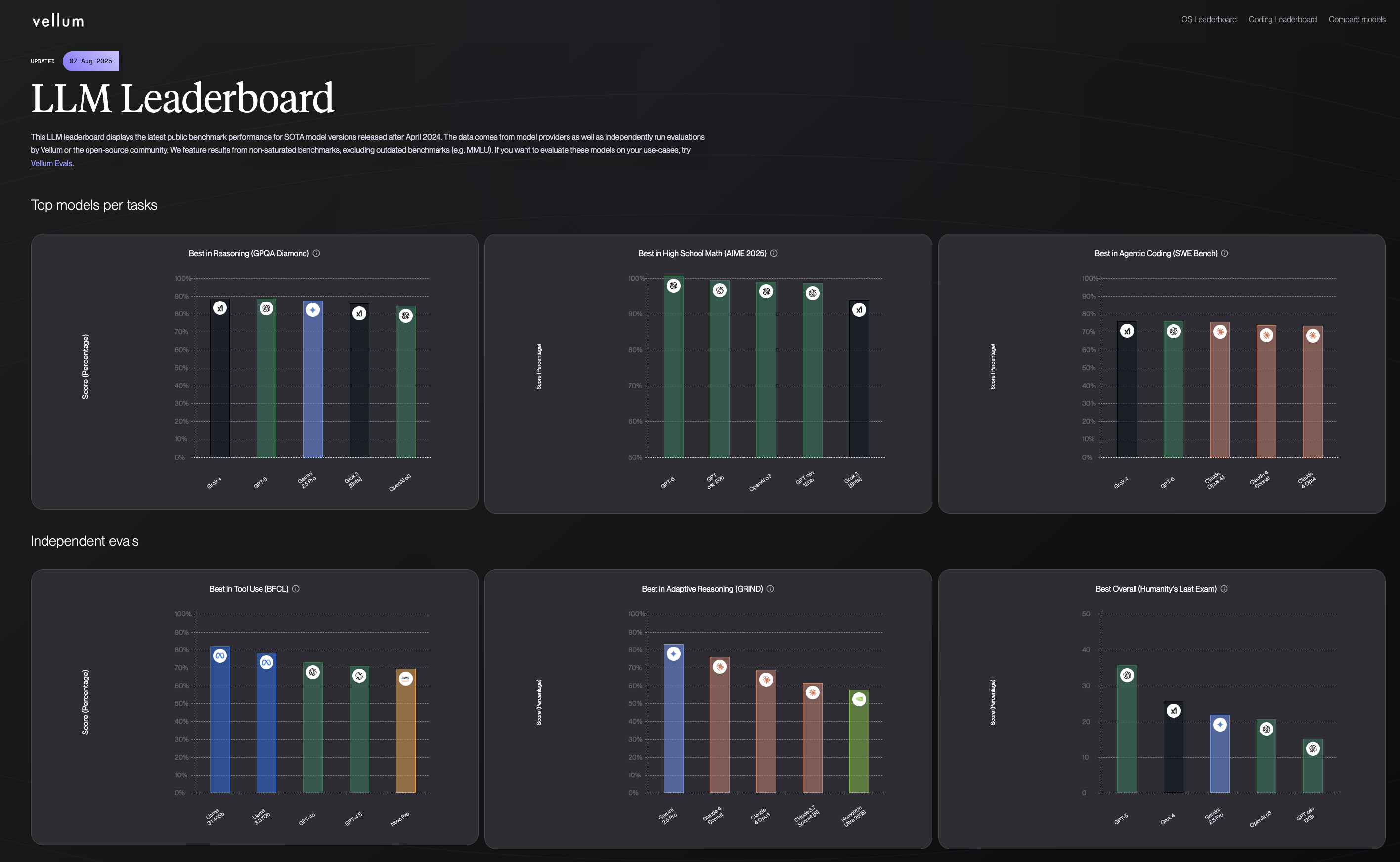}
  \caption{LLM Leaderboard illustration: Vellum leaderboard \cite{vellumLeaderboard}}
  \label{fig:leader-board-vellum}
\end{figure}

Results from these suites are surfaced on widely watched leaderboards. 
The HuggingFace \emph{Open LLM Leaderboard} aggregates automatic metrics (e.g., MMLU, ARC, HellaSwag) for open-weight models. 
The LMSYS/LMArena \emph{Chatbot Arena} leaderboard ranks dialogue quality via blind, pairwise human preferences \cite{lmarenaLeaderboard}. Task-specific evaluations such as MT-Bench\,\cite{zheng2023mtbench} and vendor dashboards (e.g., Vellum-AI \cite{vellumLeaderboard}) emphasise application skills like tool use and code generation. 
Figure~\ref{fig:leader-board-vellum} illustrates a representative snapshot, underscoring how frequently updated scorecards have become a de facto currency for comparing models.

\paragraph{Bridge to our contribution \& Limits of general leaderboards.}
General boards miss requirements central to hospital-funding logic; \emph{NordDRG-AI-Benchmark} addresses them as follows:
\begin{enumerate}[nosep,leftmargin=1.4em]
  \item \textbf{Granularity} — DRG grouping needs \emph{deterministic control flow} and a rule trace; the \emph{Grouper} suite enforces full \texttt{drg\_logic} execution and requires (\texttt{drg\_nat}, \texttt{drg\_logic.id}) (Section~\ref{norddrg-benchmark-grouper}).
  \item \textbf{Source of truth} — answers should bind to a \emph{versioned, machine-readable} rule base; we release rule-complete \emph{definition tables} and \emph{expert manuals} with stable IDs (Sections~\ref{norddrg-tables}, \ref{norddrg-docs}).
  \item \textbf{Auditability} — correctness must include the decision primitive; gold keys and reference agents emit audit-ready traces (Sections~\ref{norddrg-benchmark-grouper}, \ref{subsec:lightweight-agents}).
  \item \textbf{Language} — minority-language descriptors and bilingual transfer are under-tested; FI and FI–EN excerpts make this testable (Sections~\ref{norddrg-fi}, \ref{norddrg-fi-en}).
  \item \textbf{Scoring} — governance-grade tasks need order-invariant \emph{set} outputs; the \emph{Logic} suite uses exact-match sets, and the \emph{Grouper} suite uses strict exact-match (Sections~\ref{norddrg-questions}, \ref{norddrg-benchmark-grouper}).
\end{enumerate}

\subsection{Related Work: DRGs and LLMs} \label{sec:related-drg-llm}
LLMs are increasingly explored as end-to-end engines for hospital-reimbursement workflows. \cite{wang2023drgllama} fine-tune LLaMA on 236 k multimodal MIMIC-IV episodes, achieving 52\% top-1 accuracy over 771 MS-DRGs—a strong indication that the model captures the CC/MCC rules traditionally hard-coded in grouper software. Complementing this, \cite{kwan2024medcoders} show that a tool-augmented GPT-4 rivals human coders in ICD assignment, implying that careful retrieval and validation can offset the need for proprietary data. Zero-shot modelling is pushed further by \cite{renc2024ethos}, whose transformer infers DRG-like resource strata directly from longitudinal health trajectories, hinting at reimbursement systems emergent from the structural landscape of encoded patient health states. 

Beyond accuracy metrics, recent studies ask whether LLMs truly understand DRG logic. \cite{wang2025drgsapphire} introduce \textit{DRG-Sapphire}, using reinforcement learning with rule-aware rewards and generating physician-validated rationales that explicitly cite grouping criteria. Interpretability is also central to \cite{hajialigol2023drgcoder}, whose multi-task transformer highlights token-level evidence so auditors can verify alignment with official rules. Structured knowledge helps as well: \cite{he2022kgmttbert} inject a clinical knowledge graph into BERT and observe the largest gains on DRGs whose definitions hinge on multi-entity interactions, suggesting genuine rule-level reasoning. Finally, generic LLMs remain competitive; \cite{boyle2023automated} repurpose them for zero-/few-shot ICD coding by exploiting the ICD hierarchy, while \cite{boukhers2024llmicd} couple LLaMA representations with a lightweight MultiResCNN, reporting F1 gains across the full label set.

\paragraph{Bridge to our contribution.}
Prior DRG–LLM studies either rely on proprietary data, cover a single national system, or simplify grouping to flat classification. We address these gaps with a public, multilingual, rule-complete suite that tests models on the full NordDRG graph and yields comparable, reproducible baselines.

\subsection{DRG Grouper Software} \label{subsec:drg-grouper-software}
\emph{DRG groupers} are rule-based software engines that map a patient episode’s minimum dataset—principal and secondary diagnoses, procedures/interventions, demographics (age/sex), discharge status and other context—onto a single payment group. While national families differ (e.g., MS-DRG in the United States, AR-DRG in Australia, APR-DRG used by many payers, and NordDRG across the Nordics), they share a governed control flow: MDC entry, ordered rule evaluation, surgical evidence (OR/non-OR), complication/comorbidity handling (CC/MCC or analogous constructs), and versioned annual updates \cite{Busse2011DRGEurope,CMS2025MSDRG,IHACPA2022ARDG,ThreeM2022APRDRG,NCCReadTables}. 

In practice, groupers are distributed as compiled binaries, libraries or services accompanied by technical manuals and release notes. MS-DRG publishes detailed definitions and versioned release documentation \cite{CMS2025MSDRG}; AR-DRG provides a public technical specification describing hierarchy, splitting rules, and episode complexity scoring \cite{IHACPA2022ARDG}; APR-DRG documents clinical logic and the severity/risk subclasses layered atop base DRGs \cite{ThreeM2022APRDRG}; and NordDRG uniquely exposes machine-readable definition tables that encode the executable rule graph (diagnosis/procedure properties, age/sex bounds, national activation) \cite{NCCReadTables}. Historically, comparative scholarship has framed these systems as transparent, auditable instruments for hospital payment and CaseMix analysis, while noting national tailoring and update governance \cite{Busse2011DRGEurope}.

\paragraph{Bridge to our contribution.}
Our benchmark operationalises this ecosystem by packaging rule-complete, machine-readable artefacts (tables and manuals) and tasks that require exact, specification-faithful execution. By evaluating models against \emph{both} the final DRG and the triggering rule row, we align with grouper software’s governed control flow and make LLM behaviour auditable in the same terms as production groupers.

\subsection{Our previous work} \label{sec:our-previous-work}
Our preliminary results appeared at \cite{pitkaranta2024chira,pitkaranta2024pcsi} .  
The current article substantially extends that work by: 
(i) releasing a fully curated benchmark with task-oriented prompts, 
(ii) providing open access github repository with all artefacts and  
(iii) demonstrating baseline performance of recent LLMs on the automatically verifiable tasks

\paragraph{Bridge to our contribution.}
Building on our earlier work, this article contributes a curated release (artefacts, prompts, answer keys), a documented repository, and first-cut baselines across multiple endpoints. These additions transform earlier proof-of-concept into an extensible benchmark aligned with Objectives O1-O3 (Section~\ref{sec:research-objectives}).


\section{Research methodology}\label{sec:benchmark-research-methodology}

This work follows the \emph{Design Science Research} (DSR) paradigm, which couples rigorous inquiry with artefact construction to solve a practical problem while contributing to knowledge. In line with the six-step model of Peffers \textit{et al.}\,\cite{peffers2020designscienceresearchprocess}, we:

\begin{enumerate}[nosep,leftmargin=1.6em]
  \item \textbf{Problem identification.} Established the absence of a public, rule-complete CaseMix benchmark as a critical application–evaluation gap (see Section~\ref{sec:problem-identification} and Section~~\ref{sec:research-gap}).
  \item \textbf{Objective definition.} Specified objectives for a reproducible LLM testbed (O1–O3; Section~\ref{sec:research-objectives}).
  \item \textbf{Design \& development.} Constructed the artefact bundle (\textbf{A1–A6}; Section~\ref{sec:design-development}) that encodes the full NordDRG rule graph and governance semantics.
  \item \textbf{Demonstration.} Instantiated two suites—\emph{Logic} and \emph{Grouper}—and exercised them through task scenarios grounded in the released tables and manuals (Sections~\ref{norddrg-questions}, \ref{norddrg-benchmark-grouper}).
  \item \textbf{Evaluation.} Measured baseline performance of representative LLM endpoints under a no-web, artefact-only setting with exact-match scoring (Section~\ref{sec:benchmark-demo}).
  \item \textbf{Communication.} Released a versioned, open repository with schema-stable identifiers, prompts, gold keys, and scoring scripts to support replication and extension (Section~\ref{sec:conclusions}).
\end{enumerate}

Two rapid design–evaluate iterations with domain experts ensured that artefacts and tasks remain aligned with real-world reimbursement workflows. All evaluations fixed the \textbf{Definition Tables + PDF Instructions} configuration and constrained models to the provided materials (no external retrieval), matching governance-grade auditability requirements.

\vspace{2pt}
\noindent\footnotesize
\textbf{Author note: LLM use} \footnote{During manuscript preparation the authors made limited use of OpenAI services (\textit{GPT-5 Thinking}, \textit{GPT-4o}, \textit{o3}, and \textit{o4-mini}) for grammar, stylistic refinement, and reference-list formatting. All prompts, intermediate outputs, and final wording were reviewed and verified by the authors, who accept full responsibility for the content.}
\normalsize

\subsection{Research gap}\label{sec:research-gap}
Recent studies probe whether large language models can predict DRGs \cite{wang2023drgllama,wang2025drgsapphire} or internalise isolated grouper rules \cite{hajialigol2023drgcoder,he2022kgmttbert}, yet the evidence base remains fragmented. Typical evaluations (i) focus on a single national system, (ii) flatten multi-table CaseMix logic into single-label classification, and (iii) rely on non-redistributable (proprietary) data. Consequently, to the best of our knowledge (August~2025), there is no \emph{public, multilingual, rule-complete} benchmark that reflects the full NordDRG rule graph—spanning cross-table joins, governance manuals, and annual change-control workflows. Without such a yardstick, researchers cannot (a) establish reproducible baselines for reasoning over reimbursement logic, (b) compare design patterns (fine-tuning, retrieval, reinforcement learning, chain-of-thought) under identical conditions, or (c) track progress across model generations and research groups (see Section~\ref{sec:research-objectives}, O1–O3).

These shortcomings crystallise into two open challenges:
\begin{enumerate}[nosep,leftmargin=1.4em]
  \item \textbf{Scope gap} — No existing evaluation covers the \emph{full}, multilingual CaseMix ruleset used in hospital funding; current DRG experiments rely on reduced or proprietary slices. \emph{(Addressed by O1 via a rule-complete artefact bundle.)}
  \item \textbf{Benchmark gap} — The absence of an open, standardised CaseMix benchmark precludes rigorous, head-to-head comparison of LLM approaches and hinders cumulative research on trustworthy automation in hospital finance. \emph{(Addressed by O2–O3 via task suites and exact-match scoring.)}
\end{enumerate}

\subsection{Research Design Objectives}\label{sec:research-objectives}

In response to the \emph{Scope Gap} and \emph{Benchmark Gap} (Section~\ref{sec:research-gap}), we pursue three objectives aligned with DSRM:

\begin{enumerate}[nosep,leftmargin=1.4em]
  \item \textbf{O1 — Artefact construction} \textit{(addresses the Scope Gap; DSRM: Design \& Development).}\,
        Release an open, machine-readable, versioned benchmark (\textit{NordDRG-AI-Benchmark}) that captures the \emph{complete} NordDRG rule graph—definition tables, diagnosis/procedure property tables, age/sex splits, national activation flags, and governance manuals—with schema-stable identifiers and modular packaging for reuse and extension (see Section~\ref{sec:design-development}).

  \item \textbf{O2 — Benchmark operationalisation} \textit{(bridges both gaps; DSRM: Demonstration).}\,
        Instantiate two complementary test suites with exact-match scoring:
        (i) the \emph{NordDRG Logic Benchmark}—13 rule-level questions reflecting coder, clinician, and policy-analyst workflows, with public answer keys for quantitative tasks and rubric guides for qualitative ones (Section~\ref{norddrg-questions}); and
        (ii) the \emph{NordDRG Grouper Benchmark}—13 case-level tasks that require faithful execution of the specification’s control flow and return both \texttt{drg\_nat} and the triggering \texttt{drg\_logic.id} (Section~\ref{norddrg-benchmark-grouper}).

  \item \textbf{O3 — Baseline evaluation} \textit{(targets the Benchmark Gap; DSRM: Evaluation).}\,
        Provide first-cut, fully reproducible results for representative state-of-the-art LLM endpoints across providers, including prompts, context bundles and model outputs — establishing the initial yardstick for head-to-head comparisons and longitudinal tracking in hospital-funding contexts (Section~\ref{sec:benchmark-demo}).
\end{enumerate}

\noindent\textbf{Alignment back to the Research Gap (Section~\ref{sec:research-gap}).}
O1 mitigates the \emph{Scope Gap} via rule-complete, multilingual artefacts; 
O2 operationalises both gaps into a standardised, governance-grade benchmark; 
O3 directly addresses the \emph{Benchmark Gap} with reproducible baselines and exact-match metrics.

\section{Design \& Development: NordDRG AI Benchmark for LLMs} \label{sec:design-development}

\textbf{NordDRG-AI-Benchmark} turns the problem framing into a modular, rule-complete testbed. It bundles the NordDRG rule base—\(\sim\!20\) interlinked definition tables, annual governance manuals, and curated FI / FI-EN table excerpts—into versioned, machine-readable artefacts (A1-A6) that can be loaded singly or in combination. Coupled with task prompts and gold answer keys for the \emph{Logic} and \emph{Grouper} suites, the framework yields a reproducible, governance-grade yardstick: models are evaluated not only on the final DRG but, where applicable, also on the triggering \texttt{drg\_logic.id}. Schema-stable identifiers support drop-in annual updates and head-to-head comparisons of prompting, retrieval augmentation, and fine-tuning strategies.

All resources are released at GitHub~\footnote{https://github.com/longshoreforrest/norddrg-ai-benchmark} and are grouped into four artefact classes:

\begin{itemize}
  \item \textbf{A1 — NordDRG Full Specification}: Section~\ref{norddrg-tables} (\texttt{.xlsx/.csv}):  
        roughly twenty relational sheets that encode the complete NordDRG
        grouper logic—diagnoses, NCSP procedures, age/sex splits,
        country-activation flags, surgery properties and grouping features.

  \item \textbf{A2 — Expert documentation} Section~\ref{norddrg-docs} (\texttt{.pdf}):  
        governance manuals, change-log templates, and column-by-column
        guides that explain how technical updates are proposed, audited,
        and rolled into the annual NordDRG release.

  \item \textbf{A3 — NordDRG Logic Benchmark}: Section~\ref{norddrg-questions} (\texttt{.xlsx}):  
        a standalone, rule-oriented suite of \textbf{13} automatically verifiable CaseMix questions 
        (\emph{Logic-1}–\emph{Logic-13}). The prompts cover table navigation and retrieval, multilingual
        terminology grounding (FI-EN), cross-table joins for maternal delivery and newborn care under explicit
        inclusion/exclusion constraints, procedure-property tracing, diagnosis-property expansion and mapping,
        and CC/MCC validity checks. Each prompt is tagged \emph{Easy}, \emph{Medium}, or \emph{Hard}. Gold
        answers are provided as order-invariant, de-duplicated code sets (or Yes/No decisions) for exact-match
        scoring.
  
  \item \textbf{A4 — NordDRG Grouper Benchmark}: Section~\ref{norddrg-benchmark-grouper} (\texttt{.xlsx}):  
        13 grouping tasks (\emph{Grouper-1}-\emph{Grouper-13}) that fully emulate the NordDRG
        specification’s grouping logic, enabling standardised, reproducible
        evaluation.  Each task is paired with a gold-standard answer sheet.

  \item \textbf{A5 — Curated table excerpts (FI / FI–EN)} Section~\ref{norddrg-fi-en} (\texttt{.xlsx/.csv}):  
        a Finnish subset released in two flavours—(i) Finnish-only labels and  
        (ii) parallel Finnish–English labels—so researchers can probe LLM
        performance on minority-language data and bilingual transfer without
        loading the entire corpus.

  \item \textbf{A6 — Lightweight reference agents.} 
        Two single-role, provider-agnostic agents for no-code use in commercial LLM platforms. Both operate strictly on the released artefacts (\emph{no web/RAG/tools}) and emit an auditable evidence trace (sheet/row IDs, predicate outcomes). \emph{LogicAgent} (Combined workbook + 2 PDFs) answers Logic-1–Logic-13 as comma-separated, order-invariant lists. \emph{GrouperAgent} (FI workbook + 2 PDFs) emulates the Finnish grouper for Grouper-1–Grouper-13, returning \verb|drg_nat=<DRG>, drg_logic.id=<ID>|. See Section~\ref{subsec:lightweight-agents} for setup (file paths and system prompts).        

\end{itemize}

\noindent
Each artefact is self-contained, enabling composition into the four canonical ``dataset configurations'' shown in Table~\ref{tab:norddrg-configurations}. Researchers can start small—with the Finnish-only or FI--EN excerpts—for quick iteration, and scale up to the full definition tables together with governance PDFs when evaluating chain-of-thought reasoning and cross-document grounding.

\paragraph{Datasets and evaluation suites}
Table~\ref{tab:norddrg-resources} summarises the released resources—definition tables, governance manuals, curated FI / FI--EN subsets, and auxiliary prompts. These artefacts are self-contained and compose into the dataset configurations in Table~\ref{tab:norddrg-configurations} (e.g., \textbf{Definition Tables only} or \textbf{Definition Tables + PDF Instructions}; lightweight FI / FI--EN excerpts for rapid prototyping). The benchmark exposes two complementary suites: \emph{NordDRG Logic} (13 rule-level tasks) and \emph{NordDRG Grouper} (13 case-level emulation tasks). Either suite can be executed under any configuration—using tables only isolates rule-graph competence, while adding PDFs tests documentation-grounded reasoning and governance semantics. Scoring is uniform across configurations: \emph{Logic} uses order-invariant set equality; \emph{Grouper} requires an exact pair \((\texttt{drg\_nat},\,\texttt{drg\_logic.id})\). The subsections that follow unpack each resource and illustrate typical combinations for downstream experiments.

\newcolumntype{L}[1]{>{\raggedright\arraybackslash}p{#1}}

\begin{table}[ht]
    \centering
    \caption{Publicly released NordDRG benchmark resources.}
    \label{tab:norddrg-resources}
    \begin{tabular}{L{4.0cm} L{2.0cm} L{6.0cm}}   
      \toprule
      \textbf{Dataset} & \textbf{Format} & \textbf{Description} \\[2pt]
      \midrule
      Definition Tables                         & XLSX & NordDRG Definition Tables \\ 
      PDF Instructions                           & PDF  & NordDRG PDF instructions \\ 
      Definition Tables: Part of the tables FI   & XLSX & NordDRG Definition Tables—Finnish DRG list with Finnish descriptions only \\ 
      Definition Tables: Part of the tables FI–EN & XLSX & NordDRG Definition Tables—Finnish DRG list with Finnish \emph{and} English descriptions of groups \\ 
      Additional Prompts                         & TXT  & NordDRG additional prompts \\ 
      \bottomrule
    \end{tabular}
\end{table}

\subsection{Artefact: NordDRG Full Specification - Definition Tables}\label{norddrg-tables}

Artefact~\ref{norddrg-tables} packages the authoritative NordDRG rule base in a
machine-readable workbook of \(\sim\!20\) inter-linked sheets. It constitutes the
\emph{full specification} (\emph{rule-complete}) of the NordDRG grouping logic as executed by
production groupers: all rule-bearing tables and columns needed for deterministic control flow are
included (e.g., \texttt{drg\_logic}, \texttt{dg\_feat}, \texttt{proc\_feat}, CC/MCC exclusions,
age/sex bounds, MDC entry, and national activation flags). The same schema is used across
\emph{annual} releases and \emph{national} variants; per-year and per-country workbooks are
distributed in an identical format, enabling drop-in updates and cross-version comparisons via
stable identifiers (e.g., \texttt{DRG}, \texttt{ICD}, \texttt{NCSP}, \texttt{ORD}).

Core tables include:

\begin{itemize}[nosep,leftmargin=1.5em]
  \item \texttt{drg\_logic} — rules that map diagnosis,
        procedure, age limits, CC/MCC status, and surgery flags to a
        unique DRG code; ordered by the \texttt{ORD} execution column.
  \item \texttt{dg\_feat} / \texttt{proc\_feat} — full ICD-10 and NCSP
        diagnosis and procedure code catalogues annotated with properties referenced by
        \texttt{drg\_logic} (e.g.\ \texttt{DGCAT}, \texttt{COMPL},
        \texttt{OR}, \texttt{PROCPRO}).
  \item Lookup sheets such as \texttt{compl\_excl},
        \texttt{procprop\_name}, and \texttt{dgprop\_name}, which define
        complication categories, OR-class mappings, and diagnosis
        property labels.
  \item \texttt{age\_split} and \texttt{country\_flag} tables that hold
        paediatric/geriatric break-points and activation flags for each
        national version.
\end{itemize}

\begin{figure}[!h]
    \centering
    \includegraphics[width=1.0\textwidth]{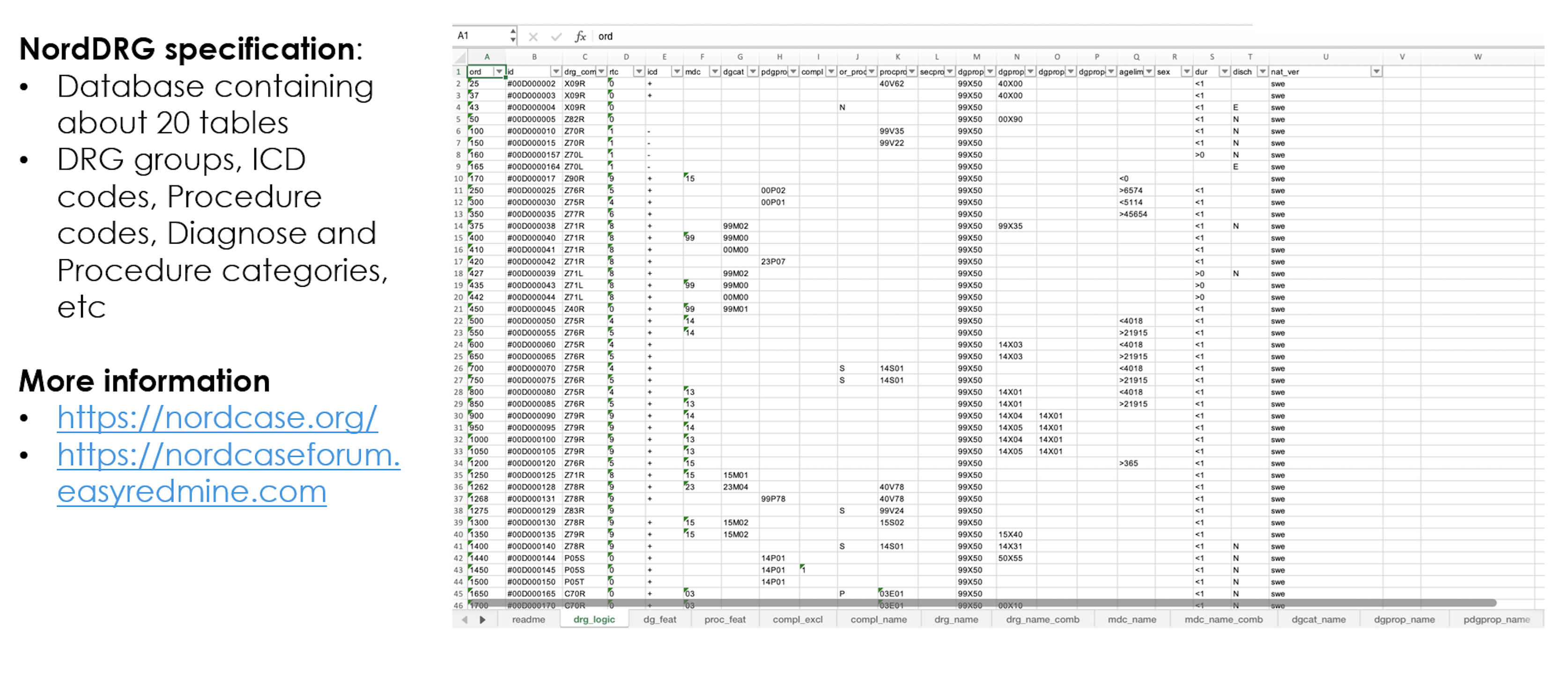}
    \caption{Materials used: NordDRG definition tables \cite{nordic_issue_tracking}}
\label{fig:materials-norddrg-definition-tables}
\end{figure}

Each worksheet exposes stable identifier columns (e.g., \texttt{DRG},
\texttt{ICD}, \texttt{NCSP}, \texttt{ORD}), allowing the tables to be
joined loss-lessly and extended by appending new rows in future
releases.  A representative excerpt is shown in
Figure~\ref{fig:materials-norddrg-definition-tables}.

\subsection{Artefact: Expert documentation}\label{norddrg-docs}

Artefact \ref{norddrg-docs} bundles the procedural knowledge that turns the raw tables
into a living, governable system.  Two official handbooks from the Nordic
Casemix Centre—supplied as searchable PDFs—and the associated Excel
template are included:

\begin{itemize}[nosep,leftmargin=1.4em]
  \item \emph{How to read the NordDRG definition tables}
        A walkthrough of the Excel workbook that
        constitutes the grouper logic.  The manual explains every
        high-impact sheet—including \texttt{drg\_logic},
        \texttt{dg\_feat}, \texttt{proc\_feat} and
        \texttt{compl\_excl}—and shows how to (i) locate all rules that
        drive a given DRG, (ii) identify which diagnosis or procedure
        properties generate CC/MCC splits, and (iii) trace OR-flags,
        MDCs, age limits, and other row-level conditions.
  \item \emph{How to write technical changes for NordDRG}
        The canonical SOP for annual updates.
        It defines the “IN/OUT” convention, file-naming rules for
        \texttt{TC\_TEMPLATE\_YYYY-MM-DD}, and per-table editing
        guidelines—for example, duplicating a \texttt{drg\_logic} row
        when altering \texttt{ORD} or \texttt{ID}, propagating new
        COMPL values consistently to \texttt{dg\_feat},
        \texttt{compl\_name}, and \texttt{compl\_excl}, and adding
        English code texts in the comments field for new ICD/NCSP codes.
\end{itemize}

\begin{figure}[!h]
    \centering
    \includegraphics[width=1.0\textwidth]{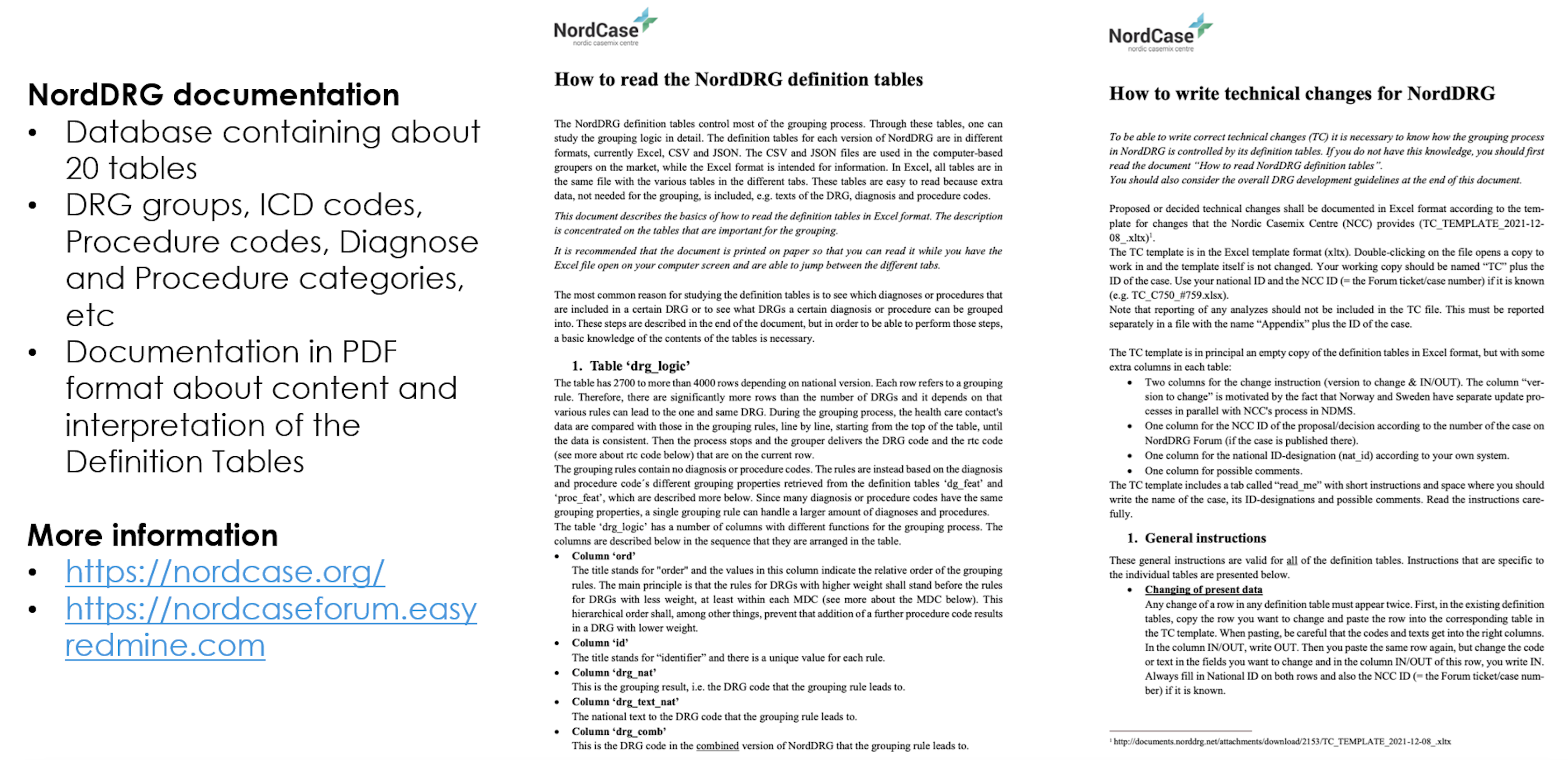}    
	\caption{Materials used: NordDRG definition table expert documentation \cite{nordic_issue_tracking}}
\label{fig:materials-norddrg-expert-documentation}
\end{figure}

Together, these documents specify (i) how existing grouping rules must be
interpreted and (ii) how proposed modifications are documented, validated
and merged.  Including them in the benchmark ensures that LLM evaluations
test not only code look-ups but also compliance with NordDRG’s governance
process.  A sample page is shown in
Figure~\ref{fig:materials-norddrg-expert-documentation}.

\subsection{Artefact: NordDRG Logic Benchmark} \label{norddrg-questions}

Artefact~\ref{norddrg-questions} provides a self-contained NordDRG Logic Benchmark focused on rule-level reasoning. It comprises \textbf{13 automatically verifiable CaseMix questions} (\emph{Logic-1}–\emph{Logic-13}) that mirror day-to-day queries posed by clinicians, health-economists, and DRG coders, while directly exercising the definition tables in Section~\ref{norddrg-tables} and, where needed, the expert documentation in Section~\ref{norddrg-docs}. In DSRM terms, this artefact \emph{operationalises} the benchmark (Objective~O2 in Section~\ref{sec:research-objectives}) and supplies machine-checkable targets that support baseline evaluation (Objective~O3 in Section~\ref{sec:research-objectives}).

\paragraph{What the questions test.}
The prompts span the main reasoning patterns required to work with NordDRG:
\begin{itemize}[leftmargin=1.4em]
  \item \textbf{Table navigation and retrieval} (\emph{Logic-1}/\emph{Logic-2}): enumerate DRG groups for a clinical theme such as appendectomy by reading \emph{drg\_logic} and group lists without duplication.
  \item \textbf{Multilingual terminology grounding} (\emph{Logic-3}/\emph{Logic-5}): resolve Finnish terms (e.g., \emph{umpilisäke}) to the correct DRG sets in Finnish and pan-Nordic variants, probing minority-language handling.
  \item \textbf{Cross-table joins over clinical events} (\emph{Logic-4}, \emph{Logic-10}, \emph{Logic-11}): list DRGs for maternal delivery and newborn care under explicit inclusion/exclusion constraints across MDCs and care settings; where definitions are ambiguous, consult the expert manuals (Section~\ref{norddrg-docs}) for interpretation.
  \item \textbf{Procedure logic} (\emph{Logic-6}): trace from a DRG to its NCSP procedure codes via procedure properties (e.g., \emph{OR}/\emph{PROCPRO}) to recover the inducing code set.
  \item \textbf{Diagnosis-property reasoning} (\emph{Logic-7}–\emph{Logic-9}): (i) link DRGs to diagnosis codes through property/category mappings, (ii) expand a property such as \emph{14X03} to the full ICD list, and (iii) read back which diagnosis categories/properties drive a specific DRG row.
  \item \textbf{Complication (CC/MCC) logic} (\emph{Logic-12}/\emph{Logic-13}): decide whether a candidate secondary diagnosis counts as a valid complication for a given principal diagnosis, reflecting exclusion lists and category rules; the manuals in Section~\ref{norddrg-docs} clarify these semantics.
\end{itemize}

\paragraph{Format and scoring.}
Table~\ref{tab:norddrg-usecases-quant} lists the 13 prompts; their targets are \emph{deterministic code sets or binary decisions}, enabling \textbf{exact-match} scoring against the gold sheet in Table~\ref{tab:norddrg-usecases-answers}. Answers are normalised as \emph{order-invariant, de-duplicated} sets of DRG or ICD/NCSP codes (or Yes/No for the complication checks). Each prompt carries a difficulty tag (\emph{Easy/Medium/Hard}) that approximates cognitive load and the number of artefacts/joins required (e.g., simple list retrieval vs.\ multi-table joins with property expansion and exclusions). This scoring design directly supports Objective~O3 (Section~\ref{sec:research-objectives}) by enabling reproducible, head-to-head comparisons.

\paragraph{Reproducibility and extensibility.}
A shared identifier (\emph{Logic-1}–\emph{Logic-13}) ties each question row to its gold answer, ensuring unambiguous evaluation. Because the prompts read only stable columns (e.g., \emph{DRG}, \emph{ORD}, \emph{MDC}, diagnosis/procedure properties) and rely on the production workbook schema, new releases can be dropped in without schema edits. Adding a task is a matter of appending one prompt row and its matching gold row; evaluation scripts can then compute exact-match accuracy automatically. Where interpretation hinges on governance conventions (e.g., exclusions or activation flags), the expert documentation (Section~\ref{norddrg-docs}) provides the authoritative reference.

Together, Table~\ref{tab:norddrg-usecases-quant} and Table~\ref{tab:norddrg-usecases-answers} form a compact, self-contained suite that surfaces whether a model can \emph{read, join, and reason over} the NordDRG rule base—complementing the full grouper emulation in Section~\ref{norddrg-benchmark-grouper} and aligning with Objectives~O2–O3 (Section~\ref{sec:research-objectives}).

\begin{table}[ht]
  \centering
  \caption{CaseMix questions with automatically verifiable answers (Logic-1--Logic-13).}
  \label{tab:norddrg-usecases-quant}
  \begin{tabular}{L{2.0cm} L{3.8cm} L{1.6cm} L{8.8cm}}
    \toprule
    \textbf{ID} & \textbf{Use Case Category} & \textbf{Difficulty} & \textbf{CaseMix Question} \\[2pt]
    \midrule
    Logic-1  & Understanding NordDRG Definition Tables    & Easy   & Could you list all DRG groups in all NordDRG national version where the main treatment during the hospital visit was related to appendectomy? Please provide a list of \texttt{drg\_comb} codes without duplicates. \\
    Logic-2  & Understanding NordDRG Definition Tables    & Easy   & Could you list all DRG groups in the Finnish DRG version where the main treatment during the hospital visit was related to appendectomy? Please provide a list of \texttt{drg\_comb} codes without duplicates. \\
    Logic-3  & Medical Terms in Local Minority Language   & Easy   & Could you list DRG groups where main treatment during the hospital visit might be related to Finnish word umpilisäke in Finnish DRG version? \\
    Logic-4  & Understanding NordDRG Definition Tables    & Medium & How would you list all NordDRG groups across all national versions where a maternal delivery (vaginal or cesarean) occurs during an inpatient, outpatient, or primary-care episode of care---excluding the 'left-over' groups (468, 470, 477, Z75R, Z76R)? \\
    Logic-5  & Medical Terms in Local Minority Language   & Medium & Could you list DRG groups where the main treatment during the hospital visit might be related to Finnish word umpilisäke in all DRG versions from all countries? Use the \texttt{drg\_comb} column codes and combine them into one list. \\
    Logic-6  & Surgical Procedure Logic                   & Medium & For DRG 166N which procedure codes have the connection through the procedure property? Use NCSP plus codes in your answer. \\
    Logic-7  & Connection between Definition Tables       & Medium & Which diagnosis codes are linked to DRG 166N through diagnosis properties in all DRG versions from all countries? \\
    Logic-8  & Diagnosis Properties                       & Medium & Which diagnosis (ICD) codes are connected to diagnosis property 14X03? \\
    Logic-9  & Connection between Definition Tables       & Medium & Which diagnosis categories and properties are listed in \texttt{drg\_logic} for DRG group 372C? \\
    Logic-10 & Understanding NordDRG Definition Tables    & Medium & How would you list all NordDRG groups across every national version where a maternal delivery (vaginal or cesarean) occurs during an inpatient or outpatient hospital episode of care---excluding primary-care delivery DRGs and the 'left-over' groups (468*, 470*, 477*, Z75R, Z76R)? \\
    Logic-11 & Understanding NordDRG Definition Tables    & Medium & How would you list all NordDRG groups across all national versions for newborn episodes of care (MDC 15) during an inpatient, outpatient, or primary-care setting---excluding the 'left-over' groups (468, 470, 477, Z75R, Z76R)? \\
    Logic-12 & Understanding NordDRG Complication Logic   & Medium & For a case with principal diagnosis A5990 and secondary diagnosis G8000 (Spastic cerebral palsy), is the secondary counted as a valid complication? \\
    Logic-13 & Understanding NordDRG Complication Logic   & Medium & For a case with principal diagnosis B3710 and secondary diagnosis J1590 (Bacterial pneumonia, unspecified), is the secondary counted as a valid complication? \\
    \bottomrule
  \end{tabular}
\end{table}

\begin{table}[ht]
  \centering
  \caption{Reference answers for each CaseMix question (Logic-1--Logic-13).}
  \label{tab:norddrg-usecases-answers}
  \begin{tabular}{L{2.2cm} L{12.3cm}}
    \toprule
    \textbf{ID} & \textbf{Right answer} \\[2pt]
    \midrule
    Logic-1  & 166C, 166N, 167N, 167X, 167O \\
    Logic-2  & 166N, 167X, 167O \\
    Logic-3  & 166N, 167X, 167O \\
    Logic-4  & 370B, 370C, 370M, 370X, 371O, 371X, 372B, 372C, 372M, 372X, 373O, 373X, 374X, 375X, P05S, P05T \\
    Logic-5  & 166C, 166N, 167N, 167X, 167O \\
    Logic-6  & JESA00, JESA01, JESA10, JESW96, JESW97 \\
    Logic-7  & C1810, K3520 \\
    Logic-8  & O6010, O6020, O6030, O6230, O6300, O6310, O6320, O6390, O6800, O6810, O6820, O6830, O6880, O6890, O7100, O7110, O7550, O7570, O8000, O8010, O8080, O8090, O8100, O8110, O8120, O8130, O8140, O8150, O8300, O8310, O8320, O8330, O8340, O8380, O8390, O8400, O8410, O8420, O8480, O8490, Z3700, Z3710, Z3720, Z3730, Z3740, Z3750, Z3760, Z3770, Z3790 \\
    Logic-9  & 14X03, 14X11, 14X13, 00X10 \\
    Logic-10 & 370B, 370C, 370M, 370X, 371O, 371X, 372B, 372C, 372M, 372X, 373O, 373X, 374X, 375X \\
    Logic-11 & 385A, 385B, 385C, 386N, 387N, 388A, 388B, 388C, 389A, 389B, 389C, 390X, 391O, 391X, 815O, 858T, 858V, 858W, 915O, 970Q, Q69L, Q93R, Q97R, Q99R, Q99U, Q99W \\
    Logic-12 & No \\
    Logic-13 & Yes \\
    \bottomrule
  \end{tabular}
\end{table}


\subsection{Artefact: NordDRG Grouper Benchmark} \label{norddrg-benchmark-grouper}

The Artefact~\ref{norddrg-benchmark-grouper} \emph{NordDRG Grouper Benchmark} requires \emph{full emulation} of the official NordDRG grouper: an LLM must take a structured test case and deterministically reproduce the same grouping decision that the specification would yield.
Concretely, the model must apply the right execution order in \emph{drg\_logic}, evaluate age/sex bounds, MDC entry conditions, surgical evidence (e.g., \emph{OR} / \emph{PROCPRO}), complication logic (CC/MCC via \emph{COMPL} and exclusions), and national activation flags to arrive at a single DRG for that case---and identify the exact \emph{drg\_logic.id} row responsible for the match.
The task therefore goes beyond code lookup: it tests end-to-end reasoning over the rule graph defined in the NordDRG specification tables (Artefact~\ref{norddrg-tables}) and their governance semantics (Artefact~\ref{norddrg-docs}).
Scoring is strict and exact-match: a submission is correct only if both the DRG code and the triggering rule identifier match the gold standard.

\paragraph{Tables and references.}
The benchmark instances are summarised in Table~\ref{tab:norddrg-usecases-grouper}, which lists 13 hard, case-level emulation tasks (\emph{Grouper-1}--\emph{Grouper-13}).
Each task references a concrete test case (by \emph{ID}) supplied in the accompanying test-case sheet and instructs the evaluator to return two fields: \emph{drg\_nat} (the resulting DRG) and \emph{drg\_logic.id} (the precise rule row).
Ground-truth outputs are provided in Table~\ref{tab:norddrg-grouper-answers} for automatic verification.
To solve these tasks, the LLM must integrate signals from the definition workbook---primarily \emph{drg\_logic} but also \emph{dg\_feat}, \emph{proc\_feat}, \emph{compl\_excl}, age-split, and country-flag sheets (Artefact~\ref{norddrg-tables})---and apply them in the correct priority order.
The inclusion of both prompts (Table~\ref{tab:norddrg-usecases-grouper}) and exact answers (Table~\ref{tab:norddrg-grouper-answers}) makes the benchmark fully reproducible and suitable for head-to-head evaluation of grouping fidelity.

\begin{table}[ht]
  \centering
  \caption{CaseMix questions for emulating the official NordDRG grouper (Grouper-1--Grouper-13).}
  \label{tab:norddrg-usecases-grouper}
  \begin{tabular}{L{2.0cm} L{2.6cm} L{1.2cm} L{10.0cm}}
    \toprule
    \textbf{ID} & \textbf{Use Case Category} & \textbf{Difficulty} & \textbf{CaseMix Question} \\[2pt]
    \midrule
    Grouper-1  & Emulating DRG grouper & Hard & Select the test case with ID 49483 from your tables. Under the Finnish (Finland) NordDRG grouper logic, which DRG group would this case be assigned to? What is the drg\_logic.id of the corresponding row? \\
    Grouper-2  & Emulating DRG grouper & Hard & Select the test case with ID 50184 from your tables. Under the Finnish (Finland) NordDRG grouper logic, which DRG group would this case be assigned to? What is the drg\_logic.id of the corresponding row? \\
    Grouper-3  & Emulating DRG grouper & Hard & Select the test case with ID 50166 from your tables. Under the Finnish (Finland) NordDRG grouper logic, which DRG group would this case be assigned to? What is the drg\_logic.id of the corresponding row? \\
    Grouper-4  & Emulating DRG grouper & Hard & Select the test case with ID 50629 from your tables. Under the Finnish (Finland) NordDRG grouper logic, which DRG group would this case be assigned to? What is the drg\_logic.id of the corresponding row? \\
    Grouper-5  & Emulating DRG grouper & Hard & Select the test case with ID 66704 from your tables. Under the Finnish (Finland) NordDRG grouper logic, which DRG group would this case be assigned to? What is the drg\_logic.id of the corresponding row? \\
    Grouper-6  & Emulating DRG grouper & Hard & Select the test case with ID 53273 from your tables. Under the Finnish (Finland) NordDRG grouper logic, which DRG group would this case be assigned to? What is the drg\_logic.id of the corresponding row? \\
    Grouper-7  & Emulating DRG grouper & Hard & Select the test case with ID 49486 from your tables. Under the Finnish (Finland) NordDRG grouper logic, which DRG group would this case be assigned to? What is the drg\_logic.id of the corresponding row? \\
    Grouper-8  & Emulating DRG grouper & Hard & Select the test case with ID 49571 from your tables. Under the Finnish (Finland) NordDRG grouper logic, which DRG group would this case be assigned to? What is the drg\_logic.id of the corresponding row? \\
    Grouper-9  & Emulating DRG grouper & Hard & Select the test case with ID 49597 from your tables. Under the Finnish (Finland) NordDRG grouper logic, which DRG group would this case be assigned to? What is the drg\_logic.id of the corresponding row? \\
    Grouper-10 & Emulating DRG grouper & Hard & Select the test case with ID 49606 from your tables. Under the Finnish (Finland) NordDRG grouper logic, which DRG group would this case be assigned to? What is the drg\_logic.id of the corresponding row? \\
    Grouper-11 & Emulating DRG grouper & Hard & Select the test case with ID 49968 from your tables. Under the Finnish (Finland) NordDRG grouper logic, which DRG group would this case be assigned to? What is the drg\_logic.id of the corresponding row? \\
    Grouper-12 & Emulating DRG grouper & Hard & Select the test case with ID 54051 from your tables. Under the Finnish (Finland) NordDRG grouper logic, which DRG group would this case be assigned to? What is the drg\_logic.id of the corresponding row? \\
    Grouper-13 & Emulating DRG grouper & Hard & Select the test case with ID 54096 from your tables. Under the Finnish (Finland) NordDRG grouper logic, which DRG group would this case be assigned to? What is the drg\_logic.id of the corresponding row? \\
    \bottomrule
  \end{tabular}
\end{table}

\begin{table}[ht]
  \centering
  \caption{Reference answers for each CaseMix grouper question (Grouper-1--Grouper-13).}
  \label{tab:norddrg-grouper-answers}
  \begin{tabular}{L{2.2cm} L{12.3cm}}
    \toprule
    \textbf{ID} & \textbf{Right answer} \\[2pt]
    \midrule
    Grouper-1  & drg\_nat=470E, drg\_logic\_id=000D004100 \\
    Grouper-2  & drg\_nat=373,  drg\_logic\_id=400D783000 \\
    Grouper-3  & drg\_nat=704O, drg\_logic\_id=106D120110 \\
    Grouper-4  & drg\_nat=468,  drg\_logic\_id=499D000100 \\
    Grouper-5  & drg\_nat=911O, drg\_logic\_id=111D169960 \\
    Grouper-6  & drg\_nat=384B, drg\_logic\_id=414D3710002 \\
    Grouper-7  & drg\_nat=388A, drg\_logic\_id=015D816140 \\
    Grouper-8  & drg\_nat=004,  drg\_logic\_id=401D041000 \\
    Grouper-9  & drg\_nat=477,  drg\_logic\_id=414D022100 \\
    Grouper-10 & drg\_nat=901O, drg\_logic\_id=101D089890 \\
    Grouper-11 & drg\_nat=125O, drg\_logic\_id=105D230000 \\
    Grouper-12 & drg\_nat=560A, drg\_logic\_id=400D313200 \\
    Grouper-13 & drg\_nat=291M, drg\_logic\_id=410D110101 \\
    \bottomrule
  \end{tabular}
\end{table}


\subsection{Artefact: Definition Tables: Part of the tables FI} \label{norddrg-fi}

Artefact \ref{norddrg-fi} distils the full workbook down to a single, lightweight file.  
The sheet contains only those DRG codes that are \emph{activated in the
Finnish version} and provides their official Finnish descriptions
(\texttt{DRG\_TEXT\_FI}).  Auxiliary columns preserve the DRG code,
MDC and version stamp so that rows can still be cross-referenced against the master tables.

This compact subset supports two complementary use cases:  
first, it enables minority-language evaluation by allowing  
researchers to probe how well an LLM grasps Finnish DRG terminology—  
without loading the full 20-sheet corpus—and, second, it facilitates  
rapid prototyping, providing an ultra-light file for few-shot  
prompting, in-context search, or on-device experiments where the complete  
tables would exceed context limits.

\subsection{Artefact: Definition Tables: Part of the tables FI-EN} \label{norddrg-fi-en}

Artefact \ref{norddrg-fi-en} mirrors the row set of the Finnish-only file but appends the
official English group descriptions (\texttt{DRG\_TEXT\_EN}).  The
bilingual layout lets LLMs align Finnish terminology with its English
counterpart, enabling cross-lingual retrieval, code-switching prompts,
and translation-quality checks within a single, lightweight worksheet.
Such functionality is essential for multilingual healthcare settings,
where clinicians, coders, and decision-support tools must seamlessly
toggle between local and international nomenclatures

\subsection{Artefact: Dataset configurations} \label{norddrg-dataset-configurations}

Artefact \ref{norddrg-dataset-configurations} defines four ready-made
modular configurations of the benchmark.  
Each primary artefact—definition tables, governance PDFs, and curated
table excerpts—can be loaded on its own or combined, depending on the
experiment.  
Table~\ref{tab:norddrg-configurations} lists four ready-made
“dataset configurations’’ that map onto typical research needs.  

The first pair, \emph{Definition Tables} and  
\emph{Definition Tables + PDF Instructions}, delivers the full-fidelity
reference material used by hospital coders and is best suited for
testing an LLM’s ability to reproduce the official NordDRG grouper logic
or to justify decisions with citations from the manuals.  

The second pair contains only a \emph{slice} of the master tables:
Finnish-only rows (FI) or bilingual Finnish–English rows (FI–EN).  
These lightweight subsets enable rapid prototyping and few-shot
prompting in resource-constrained settings while still preserving
clinically meaningful group descriptions.  

Researchers can therefore start with the compact excerpts to establish
baselines, then scale up to the full tables plus manuals to stress-test
chain-of-thought reasoning, multilingual transfer, and cross-document
grounding.  Collectively the four bundles offer a flexible design space
for analysing how data volume, document modality, and language coverage
affect LLMs capability to comprehend NordDRG classification.

\begin{table}[ht]
    \centering
    \caption{Configuration types for the NordDRG benchmark datasets.}
    \label{tab:norddrg-configurations}
    \begin{tabular}{L{4cm} L{3cm} L{4cm}}  
      \toprule
      \textbf{Dataset Configurations} & \textbf{Type} & \textbf{Description} \\[2pt]
      \midrule
      Definition Tables & Tables & NordDRG Definition Tables \\
      Definition Tables + PDF Instructions & Tables+Instructions & Definition tables and PDF descriptions \\
      Definition Tables: Part of the tables FI & PortionOfTables & NordDRG Definition Tables: Finnish DRG list with Finnish descriptions only \\
      Definition Tables: Part of the tables FI–EN & PortionOfTables & NordDRG Definition Tables: Finnish DRG list with Finnish and English descriptions of groups \\
      \bottomrule
    \end{tabular}
\end{table}

\subsection{Lightweight Agents for CaseMix Tasks}\label{subsec:lightweight-agents}
We release two \emph{single-role} agents that can be set up in minutes inside OpenAI GPTs/Assistants or Anthropic Projects/Workflows. Both operate strictly on the attached artefacts - no web browsing, RAG, function-calling, code interpreters, databases, or connectors. The only platform requirements are: (i) the ability to upload files as context and (ii) a sufficient context window to ingest one Excel workbook and two PDFs.

\paragraph{LogicAgent (provider-agnostic, no-code).}
Purpose: answer the 13 \emph{Logic} tasks deterministically from the source materials. 
Context: the \emph{NordDRG Combined tables} workbook (all national versions) and the two governance PDFs. 
\textbf{Output:} comma-separated, order-invariant, de-duplicated code lists; 
for \emph{binary} tasks (Logic-12/13) return exactly \texttt{Yes} or \texttt{No}.
Evidence: on request, the agent returns a concise \emph{evidence trace} (e.g., relevant sheet/row IDs and predicate outcomes)—not hidden chain-of-thought.

\emph{Context files (repo-root relative):}
\begin{itemize}[nosep,leftmargin=1.4em]
  \item \texttt{NordDRG\_Documentation/How\_to\_read\_NordDRG\_definition\_tables\_2021-12-20.pdf}
  \item \texttt{NordDRG\_Documentation/How\_to\_write\_technical\_changes\_for\_NordDRG\_2021-12-21.pdf}
  \item \texttt{NordDRG\_Combined2024PL/2023-06-14\_Combined2024PL\_xls.xlsx}
\end{itemize}

\emph{System prompt (verbatim):}
\begin{quote}\footnotesize\ttfamily
Read the attached tables and PDF documents. Answer the following questions only based on the documents provided to you in this context. Do not use web searches.\\
\\
Return results as comma-separated, order-invariant lists. Example: A, B, C
\end{quote}

\paragraph{GrouperAgent (Finland; governed emulation).}
Purpose: emulate the \emph{official} NordDRG grouper for the 13 \emph{Grouper} benchmark cases in the Finnish national version. 
Context: the FI definition-tables workbook and the two governance PDFs. 
Execution: applies the governed control flow over \texttt{drg\_logic}—\texttt{ORD} priority, age/sex bounds, MDC entry, OR/PROCPRO, CC/MCC (with exclusions), and FI national activation. 
Output: exactly one line \texttt{drg\_nat=<DRG>, drg\_logic\_id=<ID>}. 
Limitation: because the workbook is FI-only, cross-country Logic tasks should be run with \emph{LogicAgent} and the Combined tables.

\emph{Context files (repo-root relative):}
\begin{itemize}[nosep,leftmargin=1.4em]
  \item \texttt{NordDRG\_Documentation/How\_to\_read\_NordDRG\_definition\_tables\_2021-12-20.pdf}
  \item \texttt{NordDRG\_Documentation/How\_to\_write\_technical\_changes\_for\_NordDRG\_2021-12-21.pdf}
  \item \texttt{NordDRG\_FIN2026PL0/2024-04-15\_FIN2026PL0\_xlsx\_wo\_drgs.xlsx}
\end{itemize}

\emph{System prompt (verbatim):}
\begin{quote}\footnotesize\ttfamily
Read attached tables and PDF documents. Answer the following questions only based on the documents provided to you in this context. You are not allowed to use web searches.\\
\\
Return exactly one line with two fields:\\
drg\_nat=<DRG>, drg\_logic\_id=<ID>
\end{quote}

\paragraph{Quick start (OpenAI / Anthropic).}
\begin{enumerate}[nosep,leftmargin=1.4em]
  \item Create a tool-free agent (OpenAI GPT/Assistant or Anthropic Project/Workflow).
  \item Upload the three \emph{context files}. If the UI respects paths, keep the repo-root relative paths shown above; otherwise filenames suffice.
  \item Paste the corresponding \emph{System prompt}.
  \item \textbf{Disable} web/search/RAG and any external tools.
  \item Run Logic-1–13 with \emph{LogicAgent} or Grouper-1–13 with \emph{GrouperAgent}.
\end{enumerate}



\section{Results and Evaluation} \label{sec:benchmark-demo}

This section reports empirical findings from applying the benchmark artefacts to representative LLM endpoints. We begin by analysing baseline performance on the \emph{Logic Benchmark} (Section~\ref{norddrg-benchmark-demo-logic}) and the more demanding \emph{Grouper Benchmark} (Section~\ref{subsec:norddrg-benchmark-demo-grouper}), before considering operational throughput under consumer subscription constraints (Section~\ref{subsec:consumer-subscriptions}). We then assess how the artefacts and tasks satisfy the three design objectives O1--O3 (Section~\ref{subsec:evaluation}), linking descriptive coverage with analytical results. Taken together, these subsections provide both quantitative baselines and qualitative insights into the benchmark’s fidelity, extensibility, and governance-grade differentiation.  

\subsection{Baseline LLM Performance: NordDRG Logic Benchmark} \label{norddrg-benchmark-demo-logic}

\noindent\textbf{Evaluation setup.}
All runs used the \emph{Definition Tables + PDF Instructions} bundle
(Artefact~\ref{norddrg-tables} and Artefact~\ref{norddrg-docs}; cf.\ Table~\ref{tab:norddrg-configurations}),
and answers were constrained to the provided materials with the preamble:
``Read attached tables and PDF documents. Answer following questions only based on the documents provided for you in this context. You are not allowed to use web searches.'' 
Logic tasks are drawn from Artefact~\ref{norddrg-questions}; grouper tasks from Artefact~\ref{norddrg-benchmark-grouper}. 
We evaluated commonly used commercial endpoints from three families:
OpenAI (\emph{GPT-5 Thinking}, \emph{GPT-5 Thinking Mini}, \emph{GPT-5 Fast}, \emph{o3}, \emph{o4-mini}, \emph{4o}, \emph{4.1}),
Anthropic (\emph{Opus 4.1}, \emph{Sonnet 4}), and Google (\emph{Gemini 2.5 Pro}, \emph{Gemini 2.5 Flash}).

\paragraph{Interpreting Table~\ref{tab:norddrg-demo-results} (Logic tasks; Artefact~\ref{norddrg-questions}).}
The 13 automatically verifiable \emph{Logic-1}--\emph{Logic-13} questions separate the models into three tiers.
\emph{GPT-5 Thinking} and \emph{Opus 4.1} achieve perfect accuracy (13/13), with \emph{o3} close behind (12/13). 
A mid-tier (\emph{GPT-5 Thinking Mini}, \emph{o4-mini}, \emph{GPT-5 Fast}) lands between 6--8 correct, while \emph{4o}, \emph{4.1}, \emph{Sonnet 4}, and the Gemini models trail at 5/13 or below.
The distribution aligns with task structure defined in Artefact~\ref{norddrg-questions}:
single-table lookups or straightforward filters (e.g., \emph{Logic-1}--\emph{Logic-3}, \emph{Logic-8}, \emph{Logic-11}) are broadly solvable by the strongest systems, and \emph{Logic-12} (a CC/MCC validity check) is solved by all models.
In contrast, prompts requiring cross-sheet chaining over Artefact~\ref{norddrg-tables}---notably procedure-property tracing (\emph{Logic-6}) and diagnosis-property linkage (\emph{Logic-7})---create the largest separation.
Failures cluster around (i) omissions of relevant sheets (e.g., \textnormal{\texttt{compl\_excl}}, \textnormal{\texttt{proc\_feat}}, \textnormal{\texttt{dg\_feat}}), (ii) mishandling of ``left-over'' group exclusions, and (iii) non-deduplicated code lists that miss \emph{exact-match} scoring (as specified in Artefact~\ref{norddrg-questions}).
Overall, Table~\ref{tab:norddrg-demo-results} shows that generic leaderboard strength does not guarantee mastery of NordDRG’s property logic and governance-aware filters; however, high-end models can execute these joins reliably when given the combined artefact bundle (Artefacts~\ref{norddrg-tables}, \ref{norddrg-docs}).


\newcommand{\mh}[1]{\textbf{\small #1}}

\begin{table}[ht]
  \centering
  \caption{Accuracy on the 13 \emph{quantitative} NordDRG Logic tasks (Logic-1–Logic-13).
  All runs used the \textbf{Definition Tables + PDF Instructions} bundle. (\cmark=correct, \xmark=incorrect)}
  \label{tab:norddrg-demo-results}

  \begin{tabularx}{\textwidth}{L{1.8cm} *{11}{>{\centering\arraybackslash}X}}
    \toprule
    \textbf{Question} & \mh{GPT-5 Fast} & \mh{GPT-5 Thinking Mini} & \mh{GPT-5 Thinking} & \mh{o3} & \mh{o4-mini} & \mh{4o} & \mh{4.1} & \mh{Sonnet 4} & \mh{Opus 4.1} & \mh{Gemini 2.5 Pro} & \mh{Gemini 2.5 Flash} \\[2pt]
    \midrule
    Logic-1  & \xmark & \cmark & \cmark & \cmark & \cmark & \xmark & \xmark & \xmark & \cmark & \xmark & \xmark \\
    Logic-2  & \cmark & \cmark & \cmark & \cmark & \cmark & \xmark & \cmark & \xmark & \cmark & \xmark & \xmark \\
    Logic-3  & \cmark & \cmark & \cmark & \cmark & \cmark & \xmark & \cmark & \xmark & \cmark & \xmark & \xmark \\
    Logic-4  & \xmark & \xmark & \cmark & \cmark & \xmark & \cmark & \cmark & \xmark & \cmark & \xmark & \xmark \\
    Logic-5  & \xmark & \cmark & \cmark & \cmark & \cmark & \xmark & \xmark & \xmark & \cmark & \xmark & \xmark \\
    Logic-6  & \xmark & \xmark & \cmark & \cmark & \xmark & \xmark & \xmark & \xmark & \cmark & \xmark & \xmark \\
    Logic-7  & \xmark & \cmark & \cmark & \cmark & \xmark & \xmark & \xmark & \xmark & \cmark & \xmark & \xmark \\
    Logic-8  & \cmark & \cmark & \cmark & \cmark & \xmark & \xmark & \xmark & \xmark & \cmark & \xmark & \xmark \\
    Logic-9  & \cmark & \xmark & \cmark & \cmark & \cmark & \xmark & \xmark & \xmark & \cmark & \xmark & \xmark \\
    Logic-10 & \xmark & \xmark & \cmark & \xmark & \cmark & \cmark & \cmark & \xmark & \cmark & \xmark & \xmark \\
    Logic-11 & \cmark & \xmark & \cmark & \cmark & \cmark & \cmark & \xmark & \cmark & \cmark & \xmark & \xmark \\
    Logic-12 & \cmark & \cmark & \cmark & \cmark & \cmark & \cmark & \cmark & \cmark & \cmark & \cmark & \cmark \\
    Logic-13 & \xmark & \cmark & \cmark & \cmark & \xmark & \cmark & \xmark & \cmark & \cmark & \cmark & \xmark \\
    \midrule
    \textbf{SUM} & \textbf{6} & \textbf{8} & \textbf{13} & \textbf{12} & \textbf{8} & \textbf{5} & \textbf{5} & \textbf{3} & \textbf{13} & \textbf{2} & \textbf{1} \\
    \bottomrule
  \end{tabularx}
\end{table}


\subsection{Baseline LLM Performance: NordDRG Grouper Benchmark} \label{subsec:norddrg-benchmark-demo-grouper}

\paragraph{Interpreting Table~\ref{tab:norddrg-grouper-results} (Grouper emulation; Artefact~\ref{norddrg-benchmark-grouper}).}
\textit{Evaluation setup.} All runs used the \textit{Definition Tables + PDF Instructions} bundle (Artefacts~\ref{norddrg-tables}, \ref{norddrg-docs}), and models were explicitly instructed: “Read attached tables and PDF documents. Answer following questions only based on the documents provided for you in this context. You are not allowed to use web searches.” We evaluated commonly used commercial endpoints from three families—OpenAI (\emph{GPT-5 Thinking}, \emph{GPT-5 Thinking Mini}, \emph{GPT-5 Fast}, \emph{o3}, \emph{o4-mini}, \emph{4o}, \emph{4.1}), Anthropic (\emph{Opus 4.1}, \emph{Sonnet 4}), and Google (\emph{Gemini 2.5 Pro}, \emph{Gemini 2.5 Flash}).

\noindent The thirteen \emph{Grouper-1}--\emph{Grouper-13} tasks enforce governance-grade fidelity: an answer is correct only if it reproduces \emph{both} the final DRG and the exact triggering \textnormal{\texttt{drg\_logic.id}} defined in the specification tables (Artefact~\ref{norddrg-tables}) and required by the benchmark protocol (Artefact~\ref{norddrg-benchmark-grouper}). Under this exact-match criterion, \emph{GPT-5 Thinking} solves \textbf{7/13} cases, \emph{o3} \textbf{6/13}, and \emph{o4-mini} \textbf{3/13}; \emph{GPT-5 Thinking Mini} solves \textbf{1/13}; all remaining endpoints (\emph{GPT-5 Fast}, \emph{4o}, \emph{4.1}, \emph{Sonnet 4}, \emph{Opus 4.1}, \emph{Gemini 2.5 Pro}, \emph{Gemini 2.5 Flash}) score \textbf{0/13}. Notably, \emph{Grouper-3}--\emph{Grouper-5}, \emph{Grouper-7}, and \emph{Grouper-10} defeat all models, underscoring the gap between listing plausible DRGs and \emph{faithfully executing} the grouper’s priority order and guard rails.

\noindent The drop from \emph{Logic} to \emph{Grouper} performance pinpoints the hard parts of full specification emulation over Artefact~\ref{norddrg-tables}. Beyond table reading and joins, faithful grouping requires (i) respecting the \textnormal{\texttt{ORD}} execution priority in \textnormal{\texttt{drg\_logic}}, (ii) enforcing age/sex bounds and MDC entry criteria, (iii) integrating surgical evidence (\textnormal{\texttt{OR}}/\textnormal{\texttt{PROCPRO}}) and CC/MCC handling (via \textnormal{\texttt{COMPL}} with exclusions), (iv) applying national activation flags, and (v) mapping the decision back to the precise rule row. As DSR \emph{results}, Table~\ref{tab:norddrg-grouper-results} provides a governance-grade barometer: the best models can read and join the definition tables reliably, yet exact, traceable emulation of the governed control flow remains challenging—especially where \textnormal{\texttt{ORD}} priority interacts with CC/MCC exclusions. These errors are diagnostic rather than cosmetic: they expose divergences from the specification’s semantics (O2) and delimit the present ceiling of artefact-only emulation (O3), motivating the benchmark's insistence on reporting both \textnormal{\texttt{drg\_nat}} and \textnormal{\texttt{drg\_logic.id}}.

\newcolumntype{L}[1]{>{\raggedright\arraybackslash}p{#1}}

\begin{table}[ht]
  \centering
  \caption{DRG grouper emulation results on thirteen case-level tasks (Grouper-1–Grouper-13).
  All runs used the \textbf{Definition Tables + PDF Instructions} bundle and the structured test cases.
  (\cmark=correct; exact match on \textit{DRG} \emph{and} \textit{drg\_logic.id}; \xmark=incorrect)}
  \label{tab:norddrg-grouper-results}

  \begin{tabularx}{\textwidth}{L{2.1cm} *{11}{>{\centering\arraybackslash}X}}
    \toprule
    \textbf{Question} & \textbf{GPT-5 Fast} & \textbf{GPT-5 Thinking Mini} & \textbf{GPT-5 Thinking} & \textbf{o3} & \textbf{o4-mini} & \textbf{4o} & \textbf{4.1} & \textbf{Sonnet 4} & \textbf{Opus 4.1} & \textbf{Gemini 2.5 Pro} & \textbf{Gemini 2.5 Flash} \\[2pt]
    \midrule
    Grouper-1  & \xmark & \xmark & \cmark & \xmark & \xmark & \xmark & \xmark & \xmark & \xmark & \xmark & \xmark \\
    Grouper-2  & \xmark & \cmark & \cmark & \cmark & \cmark & \xmark & \xmark & \xmark & \xmark & \xmark & \xmark \\
    Grouper-3  & \xmark & \xmark & \xmark & \xmark & \xmark & \xmark & \xmark & \xmark & \xmark & \xmark & \xmark \\
    Grouper-4  & \xmark & \xmark & \xmark & \xmark & \xmark & \xmark & \xmark & \xmark & \xmark & \xmark & \xmark \\
    Grouper-5  & \xmark & \xmark & \xmark & \xmark & \xmark & \xmark & \xmark & \xmark & \xmark & \xmark & \xmark \\
    Grouper-6  & \xmark & \xmark & \cmark & \cmark & \xmark & \xmark & \xmark & \xmark & \xmark & \xmark & \xmark \\
    Grouper-7  & \xmark & \xmark & \xmark & \xmark & \xmark & \xmark & \xmark & \xmark & \xmark & \xmark & \xmark \\
    Grouper-8  & \xmark & \xmark & \cmark & \cmark & \cmark & \xmark & \xmark & \xmark & \xmark & \xmark & \xmark \\
    Grouper-9  & \xmark & \xmark & \cmark & \cmark & \xmark & \xmark & \xmark & \xmark & \xmark & \xmark & \xmark \\
    Grouper-10 & \xmark & \xmark & \xmark & \xmark & \xmark & \xmark & \xmark & \xmark & \xmark & \xmark & \xmark \\
    Grouper-11 & \xmark & \xmark & \cmark & \xmark & \cmark & \xmark & \xmark & \xmark & \xmark & \xmark & \xmark \\
    Grouper-12 & \xmark & \xmark & \cmark & \cmark & \xmark & \xmark & \xmark & \xmark & \xmark & \xmark & \xmark \\
    Grouper-13 & \xmark & \xmark & \xmark & \cmark & \xmark & \xmark & \xmark & \xmark & \xmark & \xmark & \xmark \\
    \midrule
    \textbf{SUM} & \textbf{0} & \textbf{1} & \textbf{7} & \textbf{6} & \textbf{3} & \textbf{0} & \textbf{0} & \textbf{0} & \textbf{0} & \textbf{0} & \textbf{0} \\
    \bottomrule
  \end{tabularx}
\end{table}


\subsection{Operational throughput under consumer subscriptions}\label{subsec:consumer-subscriptions}

\noindent\textbf{Setup.}
All evaluations were executed using standard, consumer-grade LLM subscriptions rather than enterprise contracts, to reflect the experience of typical academic and practitioner users.

\medskip
\noindent\textbf{Observation.}
Under these conditions, the \emph{Anthropic} endpoints (e.g., \emph{Sonnet~4}, \emph{Opus~4.1}) could not complete the \emph{Logic Benchmark} (Logic-1--Logic-13) in a single session: credits were exhausted mid-run, necessitating batching across multiple windows until quotas refreshed. In practice, this stretched a single end-to-end evaluation over \emph{a couple of days} due to waiting times. By contrast, the same benchmark completed without interruption on \emph{OpenAI} and \emph{Google} endpoints under otherwise identical conditions (prompt packs, context bundles, and scoring).

\medskip
\noindent\textbf{Implications for benchmarking.}
These subscription-tier effects do not change task difficulty or accuracy, but they do affect \emph{throughput} (wall-clock time to produce a full result set), \emph{operational latency} (need to pause and resume), and \emph{experiment management} (checkpointing, batching, and run logs).

\paragraph{Operational note (agents).}
A thin wrapper in the style of our \emph{LogicAgent}/\emph{GrouperAgent} (cf.\ Section~\ref{subsec:lightweight-agents}) could batch tasks deterministically, cache intermediate joins, and resume after quota interruptions—improving throughput without affecting accuracy metrics.

\medskip
\noindent\textbf{Price--value.}
Per-token price is only part of the calculus; time-to-result and quota predictability also determine value. In our runs, consumer-tier credit limits introduced delays and operator overhead (pauses, batching, etc) that can outweigh lower unit prices in time-sensitive studies. When latency is non-critical, consumer tiers may remain cost-effective. We therefore report accuracy \emph{and} operational notes to inform cost--latency trade-offs.\footnote{Subscription quotas and refill policies are product- and time-dependent; these observations reflect our August~2025 setup and may evolve.}


\subsection{Evaluation against Objectives O1--O3}\label{subsec:evaluation}

Consistent with the DSRM, we assess the benchmark against the objectives in Section~\ref{sec:research-objectives} via (i) a \emph{descriptive} inspection of the artefacts and (ii) an \emph{analytical} baseline run (Tables~\ref{tab:norddrg-demo-results} and \ref{tab:norddrg-grouper-results}).

\paragraph{O1 --- Coverage and fidelity.}
Artefact~\ref{norddrg-tables} mirrors the production NordDRG workbook (\(\sim\)20 interlinked sheets; \texttt{drg\_logic}, \texttt{dg\_feat}, \texttt{proc\_feat}, CC/MCC exclusions, age/sex splits, national activation). No lossy transformations were applied; all rule-bearing columns are preserved. The Logic tasks (Table~\ref{tab:norddrg-usecases-quant}) exercise table reading and joins; the Grouper tasks (Table~\ref{tab:norddrg-usecases-grouper}) require end-to-end execution of the official control flow (including \texttt{ORD}, MDC entry, OR/PROCPRO, CC/MCC, activation flags).

\paragraph{O2 --- Extensibility and sustainability.}
Schema-stable identifiers (\texttt{DRG}, \texttt{ICD}, \texttt{NCSP}, \texttt{ORD}, \ldots) are consistent across yearly and national releases, enabling drop-in updates without code changes. In practice, adding a version amounts to placing the official workbook and manuals in a dated repository folder; existing prompts and scoring continue to work. Maintenance follows an open workflow (GitHub issues/PRs): adding a task requires one prompt row (Artefact~\ref{norddrg-questions}) and a matching gold-answer row—no source edits.

\paragraph{O3 --- Baseline differentiation.}
On the 13 \emph{Logic} tasks, models separate clearly: \emph{GPT-5 Thinking} and \emph{Opus~4.1} achieve 13/13, \emph{o3} 12/13; mid-tier endpoints score 6--8/13; others 5/13 or below (Table~\ref{tab:norddrg-demo-results}). Largest gaps occur on cross-sheet chaining (procedure- and diagnosis-property linkage). On strict \emph{Grouper} emulation, \emph{GPT-5 Thinking} reaches 7/13, \emph{o3} 6/13, \emph{o4-mini} 3/13; all others 0–1/13 (Table~\ref{tab:norddrg-grouper-results}). Many near-misses predict the correct \texttt{drg\_nat} but fail the exact \texttt{drg\_logic.id} requirement, underscoring the difficulty of reproducing deterministic control flow and traceability. To the best of our knowledge (August~2025), this is the first public report of an LLM partially emulating the complete NordDRG grouping logic with exact matches on both fields.

\medskip\noindent
Taken together—coverage and fidelity (O1), a drop-in maintenance path (O2), and differentiated, exact-match baselines (O3)—\emph{NordDRG-AI-Benchmark} provides a robust, governance-grade yardstick for future work on trustworthy automation in hospital funding.

\clearpage

\section{Conclusions and Availability}\label{sec:conclusions}
This article introduces \emph{NordDRG-AI-Benchmark}, a rule-complete, publicly released healthcare finance benchmark for LLMs comprising two parts—\emph{Logic} and \emph{Grouper} tests. The artefacts (Section~\ref{sec:design-development}) operationalise the full NordDRG rule base and the governance documentation into machine-readable tables and task prompts, enabling exact-match, governance-grade evaluation.

\begin{figure}[!h]
  \centering
  \includegraphics[width=1.0\textwidth]{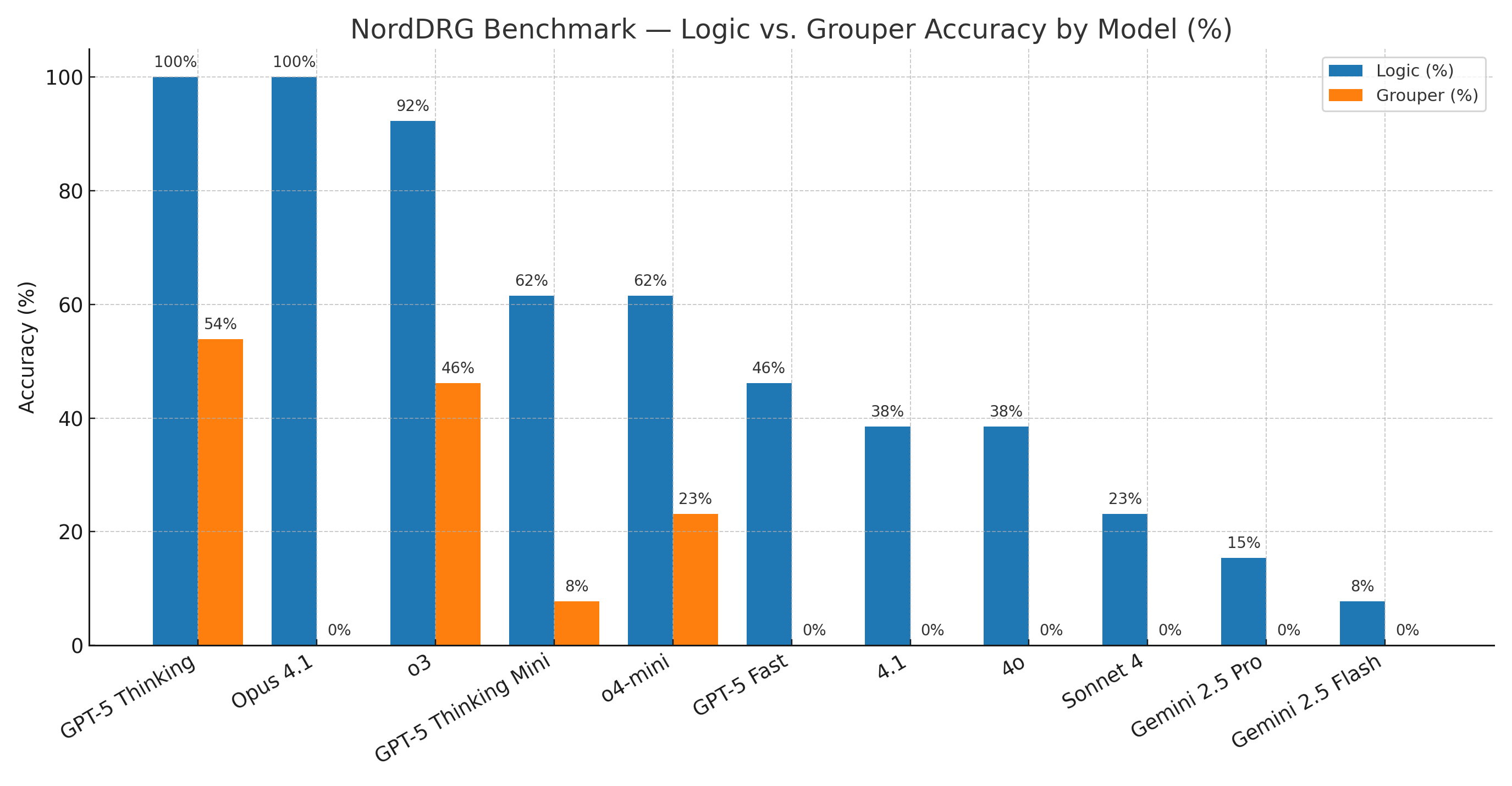}
  \caption{\textbf{Combined results.} Percentage accuracy on the \emph{NordDRG-AI-Benchmark} \emph{Logic} (rule-level) and \emph{Grouper} (full emulation) suites. Logic saturates for top models, while exact-match emulation—requiring the correct \texttt{drg\_nat} and triggering \texttt{drg\_logic.id}—remains challenging; only a few endpoints exceed \(40\%\) on Grouper. 
  \textbf{Note:} \emph{To our knowledge, this is the first public report of an LLM emulating the complete DRG grouping logic} with strict governance-grade matching.}
  \label{fig:norddrg-benchmark-results}
\end{figure}

\paragraph{Research Questions and Objectives.}
The contribution directly addresses the Research Questions (Section~\ref{sec:research-questions}) and fulfils the Design Objectives (Section~\ref{sec:research-objectives}). These findings complete the DSR cycle (Section~\ref{sec:benchmark-research-methodology}) from problem identification to evaluated artefact and public release, providing a durable, governance-grade yardstick for future work.
\begin{itemize}[nosep,leftmargin=1.6em]
  \item \textbf{RQ1\,/\,O1 (Artefact design \& construction).} Artefacts A1–A6 (Section~\ref{sec:design-development}) encode the complete NordDRG rule graph and governance semantics in an open, versioned, machine-readable form.
  \item \textbf{RQ2\,/\,O2 (Benchmark operationalisation).} Two suites—\emph{Logic} (Section~\ref{norddrg-questions}) and \emph{Grouper} (Section~\ref{norddrg-benchmark-grouper})—with exact-match scoring instantiate rule-level reasoning and full specification emulation.
  \item \textbf{RQ3\,/\,O3 (Baseline capability \& failure modes).} Baseline results (Section~\ref{sec:benchmark-demo}) and the combined chart (Figure~\ref{fig:norddrg-benchmark-results}) show ceiling effects on \emph{Logic} and partial but non-trivial success on strict \emph{Grouper} emulation.
\end{itemize}

\paragraph{Summary of results (Figure~\ref{fig:norddrg-benchmark-results}).}
\textbf{Logic (13 tasks).}
\begin{itemize}[nosep,leftmargin=1.6em]
  \item \textbf{Top tier:} \emph{GPT-5 Thinking}, \emph{Opus~4.1} — \(100\%\) (13/13); \emph{o3} — \(92\%\) (12/13).
  \item \textbf{Middle tier:} \emph{GPT-5 Thinking Mini}, \emph{o4-mini} — \(62\%\) (8/13); \emph{GPT-5 Fast} — \(46\%\) (6/13).
  \item \textbf{Trailing:} \emph{4.1}, \emph{4o} — \(38\%\) (5/13); \emph{Sonnet~4} — \(23\%\) (3/13); \emph{Gemini~2.5 Pro} — \(15\%\) (2/13); \emph{Gemini~2.5 Flash} — \(8\%\) (1/13).
\end{itemize}

\textbf{Grouper (13 tasks; strict exact-match on \texttt{drg\_nat} \& \texttt{drg\_logic.id}).}
\begin{itemize}[nosep,leftmargin=1.6em]
  \item \textbf{Best:} \emph{GPT-5 Thinking} — \(54\%\) (7/13); \emph{o3} — \(46\%\) (6/13).
  \item \textbf{Next:} \emph{o4-mini} — \(23\%\) (3/13); \emph{GPT-5 Thinking Mini} — \(8\%\) (1/13).
  \item \textbf{Others:} all remaining endpoints — \(0\%\) (0/13).
  \item \textbf{Common failure mode:} near-misses often predict the correct \texttt{drg\_nat} but a non-matching \texttt{drg\_logic.id}, which is scored as incorrect by design.
  \item \textbf{Note:} \emph{To our knowledge, this is the first public report of an LLM emulating the complete DRG grouping logic} with strict governance-grade matching.
\end{itemize}

\paragraph{Operational considerations (consumer quotas and price–value).}
In running the benchmark under standard, consumer-grade API subscriptions, we observed a material difference in \emph{throughput}: \emph{Anthropic} endpoints (e.g., \emph{Sonnet~4}, \emph{Opus~4.1}) exhausted credits mid-run and could not complete the \emph{Logic Benchmark} (Logic-1--Logic-13) in a single session, requiring batching over multiple windows with \emph{multi-day} waiting times. The same protocol completed without interruption on \emph{OpenAI} and \emph{Google} endpoints. While accuracy is unaffected, the resulting wall-clock delays and run management overhead (checkpointing, resubmission) introduce an implicit cost. Hence, the relevant objective function for practitioners is not price-per-token alone but \emph{price \& time-to-result}: for time-sensitive studies, predictable throughput may dominate nominal unit pricing.\footnote{Subscription quotas and refill policies are product- and time-dependent; these observations reflect our August~2025 setup.} (A fuller account appears in Section~\ref{subsec:consumer-subscriptions}.)

\paragraph{Availability and reproducibility.}
All artefacts, gold answers, and scoring notebooks are hosted in a public GitHub repository
(\url{https://github.com/longshoreforrest/norddrg-ai-benchmark}).
The repository follows a \texttt{folder-per-release} convention so that new national versions or yearly updates can be added by copying the official workbook and PDF manuals into a dated subfolder. The benchmark is designed for drop-in updates: prompts read schema-stable identifier columns (\texttt{DRG}, \texttt{ICD}, \texttt{NCSP}, \texttt{ORD}, etc.), and scoring relies on order-invariant, de-duplicated sets or strict exact-match keys. For full reproducibility, the repository includes: (i) prompt sheets (Logic and Grouper), (ii) gold-answer sheets, and (iii) evaluation scripts/notebooks mirroring the tables reported here.

\paragraph{Community contributions.}
Contributions are welcome via pull requests. Particularly useful additions include:
(i) new NordDRG releases or country configurations, (ii) additional prompts that exercise under-tested parts of the rule graph (with matching gold answers where deterministic), and (iii) corrections to table excerpts or documentation links. Issues can be used to propose new tasks, discuss schema nuances, or report discrepancies between Github artefacts and this documentation.

\paragraph{Limitations and future work.}
This benchmark deliberately targets \emph{specification-faithful} reasoning over the public NordDRG tables and manuals; it does not assess extraction from noisy EHR narratives or proprietary claims data. Consequently, results isolate competence on the deterministic rule graph rather than robustness to real-world documentation artifacts. 

Future work will (i) extend the Logic suite to cover remaining corners of the NordDRG specification; (ii) provide idiomatic loaders and exporters to ease integration into common toolchains; (iii) release a reproducible evaluation harness that standardises prompt packs, context assembly, version pinning, and exact-match scoring for head-to-head comparisons; and (iv) add dataset/model cards documenting governance, traceability, and known risks.

\paragraph{Closing remark.}
By packaging an auditable rule base with exact-match tasks and open scoring, \emph{NordDRG-AI-Benchmark} provides a durable yardstick for research on trustworthy automation in hospital funding—supporting head-to-head comparisons today and longitudinal tracking as models and governance requirements evolve.


\section*{Acknowledgements}
\noindent
We would like to acknowledge the \textbf{Nordic CaseMix Centre (NCC)} for its
significant contributions to the development of \emph{NordDRG-AI-Benchmark} 
by providing access to NordDRG definition tables.

We also thank \textbf{Kristiina Kahur}, CEO of NCC, for her valuable comments
on the manuscript, which helped improve the clarity and accuracy of this
paper.